\documentclass[journal]{IEEEtran}

\usepackage{xspace}

\makeatletter
\DeclareRobustCommand\onedot{\futurelet\@let@token\@onedot}
\def\@onedot{\ifx\@let@token.\else.\null\fi\xspace}
\def\eg{\emph{e.g}\onedot} 
\def\ie{\emph{i.e}\onedot} 
 
\def\etc{\emph{etc}\onedot} 
 
\def\etal{\emph{et al}\onedot}
\makeatother

\usepackage{times}
\usepackage{epsfig}
\usepackage{graphicx}
\usepackage{amsmath}
\usepackage{amssymb}
\usepackage{multirow}
\usepackage{algorithmic}
\usepackage{algorithm}
\usepackage{color}
\newcommand{\tabincell}[2]{\begin{tabular}{@{}#1@{}}#2\end{tabular}}
\usepackage{cite}
\usepackage{cite}
\usepackage{diagbox}
\usepackage{bm}
\usepackage{booktabs} 
\usepackage[pagebackref=true,breaklinks=true,colorlinks,bookmarks=false,linkcolor=blue,citecolor=blue]{hyperref}
\begin{document}
\title{Learning from Web Data: the Benefit of Unsupervised Object Localization}

\author{Xiaoxiao Sun,
	Liang Zheng,
	Yu-Kun Lai,
	Jufeng Yang
	\thanks{X. Sun and J. Yang are with College of Computer Science, Nankai University, Tianjin, 300350, China. Email:
		sunxiaoxiaozrt@163.com, yangjufeng@nankai.edu.cn}
	\thanks{L. Zheng is with Research School of Computer Science, Australian National University, Canberra, Australia. (Email: liangzheng06@gmail.com)}
	\thanks{Y.-K. Lai is with School of Computer Science and Informatics,
		Cardiff University, Wales, UK. Email: Yukun.Lai@cs.cardiff.ac.uk}
}

\markboth{IEEE TRANSACTIONS ON IMAGE PROCESSING,~2019}%
{Sun \MakeLowercase{\textit{et al.}}: Learning from Web Data: the Benefit of Unsupervised Object Localization}

\maketitle

\begin{abstract}
%Effectively training convolutional neural networks (CNNs) demands a large amount of well-labeled data, of which the annotating process is time-consuming.
Annotating a large number of training images is very time-consuming. 
In this background, this paper focuses on learning from easy-to-acquire web data and utilizes the learned model for fine-grained image classification in labeled datasets. 
Currently, the performance gain from training with web data is incremental, like a common saying ``better than nothing, but not by much''. 
Conventionally, the community looks to correcting the noisy web labels to select informative samples. 
In this work, we first systematically study the built-in gap between the web and standard datasets, \ie different data distributions between the two kinds of data.
Then, in addition to using web labels, we present an unsupervised object localization method, which provides critical insights into the object density and scale in web images.
Specifically, we design two constraints on web data to substantially reduce the difference of data distributions for the web and standard datasets.
%the built-in gap between web and standard datasets.
%
First, we present a method to control the scale, localization and number of objects in the detected region.
Second, we propose to select the regions containing objects that are consistent with the web tag.
Based on the two constraints, we are able to process web images to reduce the gap,
%between the web and
%standard datasets. 
%
and the processed web data is used to better assist the standard dataset to train CNNs.
Experiments on several fine-grained image classification datasets confirm that our method performs favorably against the state-of-the-art methods. 
\end{abstract}


\section{Introduction}
\IEEEPARstart{D}{eep} Convolutional Neural Networks (CNNs) have achieved great success in solving recognition challenges including image classification~\cite{krizhevsky2012imagenet},
scene recognition~\cite{IT3,farabet2013learning}
and fine-grained image classification~\cite{krause2016unreasonable,cui2018large,peng2018object}. 
However, the commonly used CNN models such as AlexNet~\cite{krizhevsky2012imagenet}, VGGNet~\cite{simonyan2014very}, GoogleNet-Inception~\cite{Szegedy_2015_CVPR} and ResNet~\cite{he2015deep} have a huge number of  parameters, and their performance heavily depends on the availability of a large number of labeled training examples.
In practice, getting reliably annotated images at a large scale is usually expensive and time-consuming, which prevents CNNs from being sufficiently trained for new image recognition tasks.

This paper considers the task of leveraging web data to improve recognition accuracy.
The advantage of web data is primarily in its large-scale availability, such as millions of images with user-supplied tags from social websites or image search engines~\cite{xiao2015learning,krause2016unreasonable,joulin2016learning,azadi2016auxiliary}.
\begin{figure}[!t]
\begin{center}
	%\fbox{\rule{0pt}{2in} \rule{.9\linewidth}{0pt}}
	\includegraphics[width=\linewidth]{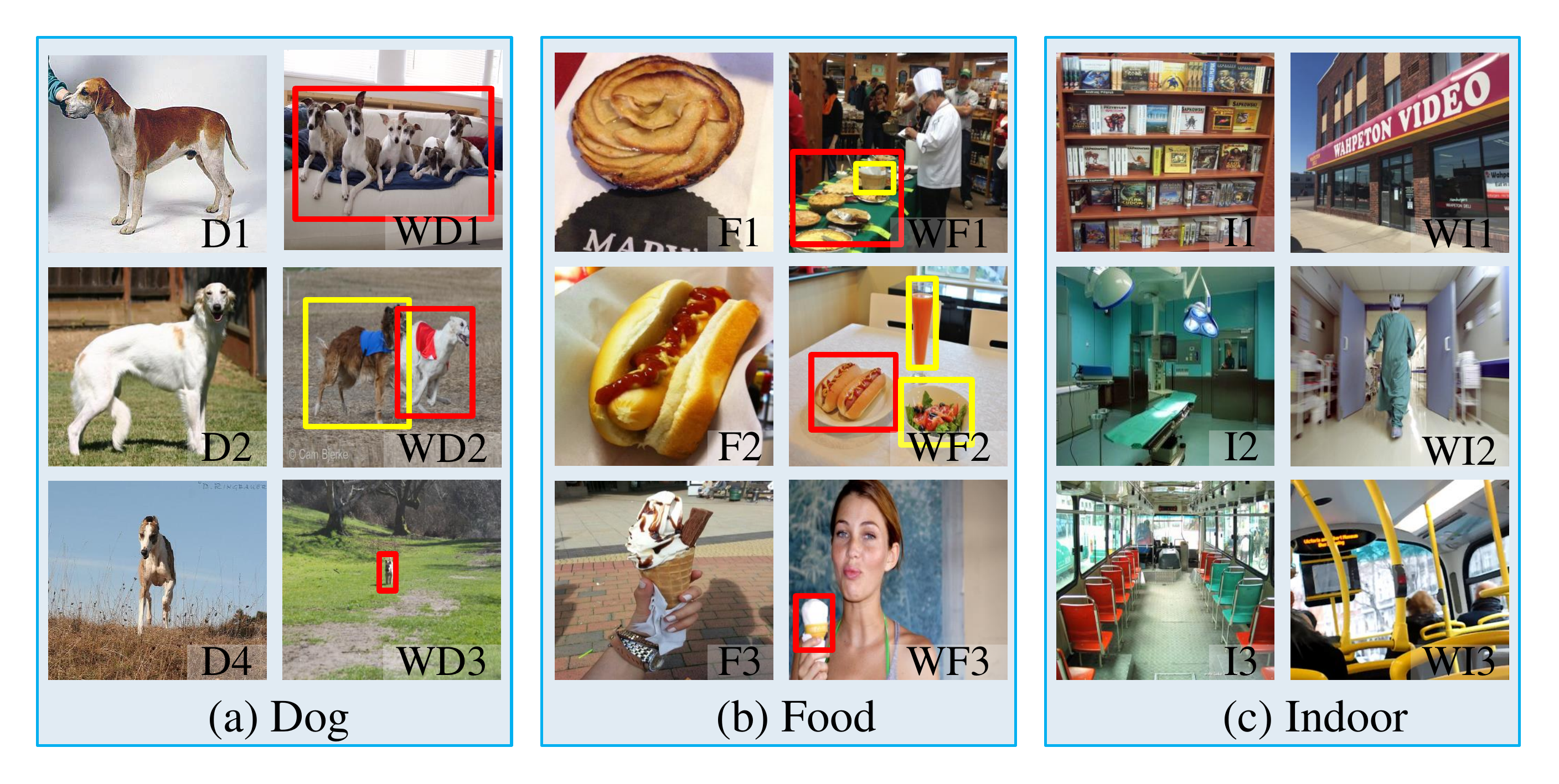}
\end{center}
	\vspace{-3mm}
\caption{Sample images of the standard and web datasets.
	%
	%The object of Dog (D), Food (F) images from standard datasets usually are  located in the center of the images.
	%
	%The object of Web images (WD, WF)
	The image pairs at each row of (a), (b) or (c) have the same label.
	The objects in Web Dog (WD-$\#$) and Web Food (WF-$\#$) images are selected by the boxes.
	Red boxes on images mean that the contained objects are consistent with the web labels, and yellow boxes denote that the objects are not consistent with the labels. 
	%%%YKL Why using different colors for showing they are different? Surely one color would be sufficient?
	%%%XXS I have modified the figure and caption.
	For (c), the images are from the MIT Indoor 67 dataset (I-$\#$) and the Web Indoor (WI-$\#$) scene images.
}
\vspace{-3mm}
\label{fig:fig1}
\end{figure}
Intuitively, more training data leads to higher classification ability of CNN models.
However, there exists a saying when employing web data, which is ``better than nothing, but not by much''. 
Existing works suggest that the labeling noise contained in webly-labeled images usually negatively influences the performance of CNN models.
However, we believe the phenomenon that the content of web images is more complex than images from standard datasets is the crux of the problem.
Fig.~\ref{fig:fig1} shows images from Food-101~\cite{bossard2014food}, Stanford Dogs~\cite{khosla2011novel}, MIT Indoor67~\cite{indoor67} datasets and web data.
The objects of Web Dog (WD-$\#$), Web Food (WF-$\#$) images are marked by bounding boxes.
Objects in red boxes are consistent with the web tag, and objects in yellow boxes are not. 
It is obvious that the objects of Dog (D1-D3) and Food (F1-F3) images from standard datasets are usually located in the center of the images and have a regular scale (size), but web images do not have these characteristics, such as WD3 and WF3.
Meanwhile, web dog and food images often contain more than one object, some of which are from different categories, such as WD2 (two breeds of dogs) and WF2 (three different kinds of food). 
On the other hand, analyzing Indoor (I1-I3) scenes requires knowledge about both scenes and various objects, but the objects in web indoor scene images are often more cluttered than standard images, \eg, WI2 and WI3.
Based on the observation, we believe that the aforementioned factors, \emph{i.e.,} the ``built-in gap'' between web and standard datasets, influence the effectiveness of utilizing web data. 
As discussed in~\cite{torralba2011unbiased}, performance on the test set of a target task often decreases when the training data is augmented with data from other datasets. 
Torralba \etal \cite{torralba2011unbiased, khosla2012undoing} argue that the reason of this phenomenon is the input space of each image dataset is different, \emph{i.e.,} the datasets are biased.
Regarding web data, we predict the label probabilities of sampled web images and test images from the Food-101 dataset using a benchmark model (fine-tuned on the Food-101 dataset with recognition accuracy of 84.31\%). 
Fig. \ref{fig:fig2} shows the predicted label probabilities for two images from the web (red) and standard (green) datasets, which are unexpectedly significantly different.
%
%The highest probabilities for dataset images correspond to correct labels, but this is not true for web images.
%
For example, the image \emph{baby-back-ribs} from Food-101 has a salient rib closer to the center of the image.
In contrast, the web \emph{baby-back-ribs} image contains four dishes, but only the one in the top left contains ribs. %
Hence, the localization and scale of the labeled content, as well as the number of relevant objects result in the difference between the data from the two sources.
To further illustrate that bias exists between web and standard datasets generally rather than occasionally,  we conduct rigorous  experiments to quantitatively analyze the difference between standard and web datasets based on 5 criteria:
relative data bias, cross-dataset generalization~\cite{torralba2011unbiased,khosla2012undoing}, the scale of related content, density of domain information, and label quality (see Section~\ref{(sec:bias)}).
%
%which influences the effectiveness of using webly-labeled data for training CNNs.
%
%
\begin{figure}[!t]
\begin{center}
	%\fbox{\rule{0pt}{2in} \rule{.9\linewidth}{0pt}}
	\includegraphics[width=\linewidth]{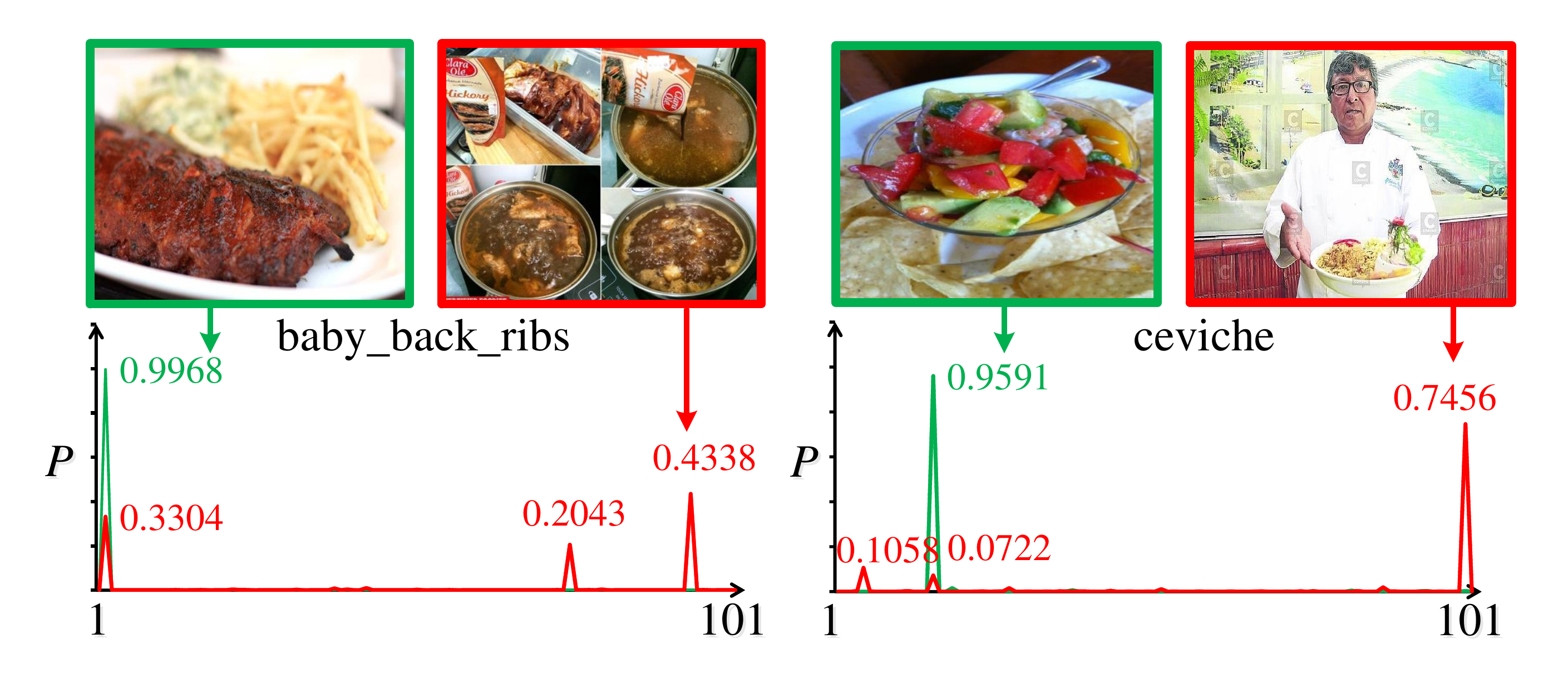}
\end{center}
	\vspace{-3mm}
\caption{Comparison of probabilities of example images from web and standard data belonging to different classes.
	Here, we show the web data (red border) and standard data (green border) from the two classes with their class names given underneath, as well as the label probabilities (\textit{P}) for 101 classes predicted by the benchmark recognition model. \label{fig:fig2} }
	\vspace{-3mm}
\end{figure}
In this paper, we design abundant experiments to evaluate the web and standard datasets, and experimental results (in Section~\ref{(sec:bias)}) confirm our conjecture that the complexity of web data results in the bias between the web and the standard datasets.
%
%In other words, web data are biased samples of a general dataset.
%JFY: 这里对三种bias的描述与第三节第一段完全不一样，全文中是否还有此类问题？必须统一！
%XXS: 杨老师我修改了这里
%JFY: 看到了你的修改，但是“in three aspects”对应哪三个？
%似乎不明确且有歧义
%XXS: 杨老师，这里改成了两点，之前三个方面中的“图像内容是否和分类相关”，“图像内容分布“中第一个和标签一致性有点重复。改成两点可能好点，会和方法里的两个约束对应。
Furthermore, we address the problem of learning from web images by reducing dataset bias in two aspects: (1) normality of content/information distribution on images (referred to as Form bias, \ie whether the location, size and number of classification objects/content on web data are similar to these conditions on standard data). (2) label consistency (referred to as Label bias, \ie whether the label is accurate to describe the web image). 
Specifically, we present an unsupervised object localization method to generate content regions related to the task on web images.
Considering standard datasets without bounding box (bbox) annotations, we employ the  ``labeled-content-centric'' prior to generate the bboxes, and use the bboxes to train a region proposal network (RPN).
Then, we generate object proposals of web images and add two constraints for the regions: 
(1) a form constraint to control the scale and density of objects based on the bbox of region proposal, and
(2) a label constraint to control the consistency of subject and the label based on the score from RPN and the probability from the benchmark recognition model.
Finally, the eligible regions are collected as the de-biased web dataset.
The evaluation results on the Food-101, Stanford Dogs (Dog-120)~\cite{khosla2011novel} and MIT Indoor 67~\cite{indoor67} datasets, as well as images retrieved from the web show that the designed framework can substantially reduce bias. 
Meanwhile, extensive experimental results show that reducing bias effectively improves the efficiency of using web data.

The contributions of this paper are three-fold.	
\begin{enumerate}
\item  We systematically study the built-in gap between web and standard datasets, and perform quantitative analysis on dataset bias from various aspects. 
As far as we know, there is no existing work that processes web data from this perspective.
\item  We propose an unsupervised object localization method and design two constraints based on the built-in gap between web and standard datasets to change the scale and density consistency of web images to approximate standard data.
\item  We perform experimental evaluation on three publicly available datasets, \emph{i.e.,} Stanford Dogs, Food-101, MIT67.
The experiments demonstrate that the proposed framework is able to improve the performance of learning from web images.
\end{enumerate}

\section{Related work}

\subsection{Deep Learning from Web Data}
Leveraging web data~\cite{zhuang2017attend,yeung2017learning,hong2017weakly,li2017learning, guo2018curriculumnet} has been an alternative method to satisfy the data requirements of deep models.
Previous work explores utilizing web data by reducing the label noise level or designing a noise-aware model~\cite{sukhbaatar2014training, xiao2015learning, krause2016unreasonable, azadi2016auxiliary}. 
Different from these works, we propose to reduce dataset bias when employing web data.

The ever-growing web images have been an important data source for computer vision tasks.
However, label noise is ubiquitous in real-world data, which influences the performance of the model.
Efforts have been made to improve results when using web data~\cite{vo2017harnessing,hu2017robust,yao2017exploiting,niu2017action,niu2016visual}.
%
%Chen \etal~\cite{chen2013neil} use a semi-supervised learning algorithm to find the relationships of common sense and labeled images of given categories. 
%%
%Schroff \etal~\cite{schroff2011harvesting} propose an automatic method for harvesting the web and gathering hundreds of images for a given query class. 
%%
%Both works above try to build visual datasets with minimum human effort, but the obtained data contains much label noise. 
%
Recent works~\cite{TDN} demonstrate that web data can be used to improve the performance of deep learning. 
To readily scale to ImageNet-sized problems, Izadinia \etal~\cite{izadinia2015deep} present a method to directly learn from image tags in the wild and show that web data is useful for training CNN models.
Chen and Gupta~\cite{chen2015webly} propose a two-stage approach to training CNNs by exploiting noisy web data and the transferability of CNNs. 
This work assumes that label noise is conditionally independent of the image, and is directly affected by the noisy annotations produced by humans.
Xiao \etal~\cite{xiao2015learning} use a probabilistic framework to handle noisy labels and train a classifier in an end-to-end learning procedure.
Furthermore, Krause \etal~\cite{krause2016unreasonable} demonstrate that using massive web data for training CNN models can improve the performance of deep models on fine-grained datasets. 
Joulin \etal~\cite{joulin2016learning} argue that ConvNets can learn from scratch in a weakly-supervised way, by utilizing 100M Flickr images annotated with noisy captions. 

However, there exists a phenomenon when employing web data for training CNN models, which is ``better than nothing, but not by much''.
%
%	Fig.~\ref{fig:fig2} shows the performance statistics of works which have used web data. 
%
In previous work, there is no significant improvement for the final classification accuracy, except for the cases which use over $10$-$100$ times more images than the standard datasets~\cite{krause2016unreasonable}.
%
%As we have shown, there is no significant improvement even after label cleaning. 
%
Different from above work, a new perspective is considered in this work and systematic experiments show that dataset bias between web and standard datasets of the target task is the main reason for the limited benefit of web data. 
Therefore, we explore the potential of web data for training CNN models by reducing bias.

\subsection{Built-in Gap between Datasets}
With the recent development of deep learning, transfer learning has been successfully applied to object recognition problems. %
Based on the observation of transferability of deep neural networks, the works~\cite{oquab2014learning,krizhevsky2012imagenet} propose to conquer this problem by first initializing CNN parameters with a model pre-trained on a larger yet related dataset, and then fine-tuning it on the smaller dataset of the target task.
However, there is a problem that the data distribution of the source task often fails to match the target task exactly (and vice versa).
We define the mismatch as dataset bias, which has been researched in~\cite{torralba2011unbiased, ECCV12_Khosla}. 
These works prove that bias exists in visual datasets, and \cite{misra2016learning} further shows that the contents of images are often biased with human-centric labels. 
Moreover,~\cite{vondrick2015learning} transfers human biases into machine systems to help with object recognition.

We believe the bottleneck of employing web data for training CNN models is the bias between web and standard datasets, so we present a method to reduce the bias for improving the performance of using web data.
In~\cite{torralba2011unbiased}, biases are summarized as selection bias, capture bias, category or label bias and negative set bias. %
\cite{misra2016learning} reveals the bias of human-centric annotations with images as keyword reporting bias.
\cite{herranz2016scene} considers the scale bias of object and scene datasets.
Based on existing summarized biases, we systematically study dataset bias between webly-labeled data and standard data.
Meanwhile, we present an unsupervised object localization method to reduce the bias, which can improve the effectiveness of using web data.

\subsection{Object Localization Methods}
\subsubsection{Traditional Object Detection Methods} 
The sliding-window method has been used for a long time in which a classifier is applied on a dense image grid.
For this kind of methods~\cite{lecun1989backpropagation,vaillant1994original}, LeCun \etal apply convolutional neural networks to handwritten digit recognition and improve the final results. 
Viola and Jones~\cite{viola2001rapid} use improved object detectors to detect faces, leading to widespread adoption of these models.
%
%The development of feature destining, such as HOG~\cite{dalal2005histograms} and integral channel features~\cite{dollar2009integral}, further improve the effectiveness of methods for pedestrian detection.
%
The introduction of Deformable Part Models (DPMs)~\cite{felzenszwalb2010cascade} helps extend dense detectors to more general object categories and get best results on the PASCAL dataset~\cite{everingham2010pascal}.
However, with the development of deep learning~\cite{krizhevsky2012imagenet}, new detectors based on deep models quickly become the main approach for object detection.
\subsubsection{Deep Object Detection Methods}
The commonly used framework in modern object detection is a two-stage approach.
Based on selective search work~\cite{uijlings2013selective}, the first stage will generate a set of candidate object proposals that contain all objects while filtering out the majority of negative locations.
Then, the second stage classifies the proposals into foreground classes and background.
On the other hand, some recent research focuses on developing one-stage methods, such as OverFeat~\cite{sermanet2014overfeat}, which is one of the first end-to-end object detectors based on deep networks, as well as SSD~\cite{liu2016ssd} and YOLO~\cite{redmon2017yolo9000}. 
These methods can improve the processing speed but their accuracy is not good compared with two-stage methods, such as R-CNN~\cite{girshick2014rich}.
%
%The design of our RetinaNet detector shares many similarities with previous detectors based on sliding-window, in particular the concept of ``anchors" introduced by RPN~\cite{ren2015faster} and utilize the features pyramids used in SSD~\cite{liu2016ssd} and FPN~\cite{lin2017feature}.
%
Different from these works, we emphasize that our simple detector achieves top results not based on innovations in network design but due to our novel strategies.
\subsubsection{Unsupervised Object Detection Methods}
Unsupervised object discovery (also called image co-localization)~\cite{cho2015unsupervised,wang2015relaxed} is a fundamental problem in computer vision, where it needs to discover the common object emerging in only positive sets of example images (without any negative examples or further supervisions).
%
%Image co-localization shares some similarities with image co-segmentation~\cite{zhao2015semantic}.
%
Recently, there also appear several co-localization	methods based on pre-trained deep convolutional models \eg, Li \etal~\cite{li2016image}.
These methods just treat pre-trained models as simple feature extractors to extract the fully connected representations, which do not sufficiently mine the information beneath the convolutional layers. 
Zhang \etal~\cite{zhuang2017attend} propose a two-level attention framework for dealing with webly-supervised classification, which achieves a new state-of-the-art.
%
%Specifically, they not only use a high-level attention focusing on a group of images for filtering out noisy images, but also employed a low-level attention for capturing the discriminative image regions on the single image level.
%
Our method handles web data based on RPN, which can (1)
recognize noisy images and also (2) supply bounding	boxes of objects.
\begin{figure*}[!t]
\begin{center}
	%\fbox{\rule{0pt}{2in} \rule{.9\linewidth}{0pt}}
	\includegraphics[width=\linewidth]{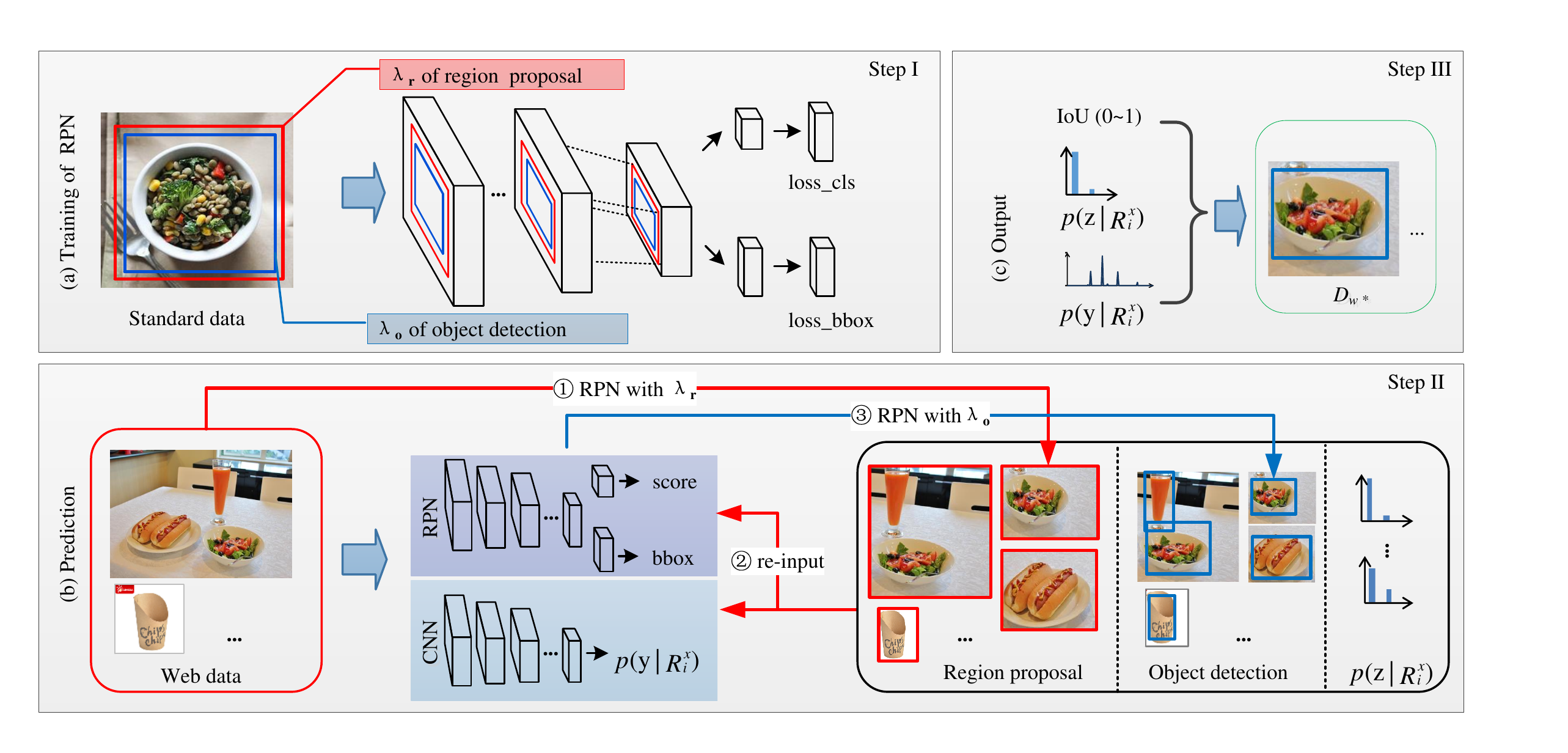}
\end{center}
\caption{Overview of the proposed method.
 (a) Training of RPN: the region proposal network is trained based on the ``labeled-content-centric'' prior.
 The unsupervised red rectangles are used to train the RPN for region detection to generate region proposals.
 Smaller blue rectangles are used to train another RPN for object detection.
%%%YKL Throughout the paper, there are two names used for these two RPNs: object proposal and object detection here, and region detection and object detection in Fig. 4. These should be made consistent.
%%% XXS have modified the two figure according to your suggestion.
 %
 This is the first step of the proposed method.
(b) Prediction:
 it is the second step to process the image and calculate the scores as well as the probabilities of the proposal regions.
 First, web data $\mathcal{D}_w$ is input into the region detection RPN (the RPN with size ratio $\lambda_{r}$ for region detection) 
%%%YKL I think the two \lambdas need to be differentiated, e.g. \lambda_{op} & \lambda_{od} or \lambda_r (for regions) and \lambda_o (for objects)
%%%XXS Thanks for the suggestion, the symbol has been modified.
 to generate region proposals $\{R^x \}$ and scores of regions $\{p(z|R_i^x)\}$. 
	Then, $\{R^x\}$ is re-input into the benchmark classification model (fine-tuned on the standard dataset) to get the predicted probabilities of labels $\{p(y|R_i^x)\}$, and is input into the RPN with $\lambda_{o}$ for object detection to generate smaller regions containing objects.
	(c) Output ${D}_{w*}$: based on the score, label probability and the approximate IoU (Intersection-over-Union), the constraints are added to control the object localization and label consistency of regions.
	The satisfactory regions are combined to form the de-biased dataset $\mathcal{D}_{w*}=\{R_{sub}^x\}$.
	Our method benefits from region proposal networks (Section~\ref{sec:GOB}) and the constraints (Section~\ref{sec:constraints}).
	\label{fig:pip}}
	\vspace{-2mm}
\end{figure*}
%
%..................................................................
\section{Method}
We aim to eliminate the dataset bias between the web and standard datasets, and then utilize such data to train a CNN model for classification.
The causes of built-in bias are summarized as two aspects in this work: Form bias and Label bias (Section~\ref{sec:type bias}).
%
%Selection bias is caused during dataset collection, which is difficult to evaluate and control before downloading the web data, so we use images from different web sources in our work to mitigate the effect of unique selection. 
%
In our method, we mainly address the form and label biases under the following assumptions: 
(1) web data may contain different distributions, such as size, number and localization of classification objects/contents from the standard dataset (e.g. classification objects/contents for the dog recognition task refer to dogs appearing in the images);
%{\color{red}(domain information means that the content of images used for food classification is related to food)}.
(2) labels of web images may be inconsistent with the contents of images.
Specifically, We have a web dataset $\mathcal{D}_w=\left \{(x^{(1)},\widetilde{y}^{(1)}),\cdots,(x^{(N)},\widetilde{y}^{(N)})\right \}$, where the $n$-th image $x^{(n)}$ has label $\widetilde{y}^{(n)} \in \left \{1,\cdots,L\right \}$, $N$ is the number of images and $L$ is the number of classes. 
Meanwhile, we have a standard dataset $\mathcal{D}_s$.
Our target is to transform web dataset $\mathcal{D}_w$ to a new dataset $\mathcal{D}_{w^*}$, which is de-biased with respect to $\mathcal{D}_s$. 

\subsection{Framework of the Presented Method}
First, we capture objects related to the target task to normalize the scale, density and location of objects, which are important measures of form bias.
Specifically, we train a detection model on the standard dataset $\mathcal{D}_s$ to detect target objects in web images, 
and then learn the level of relevance of detected regions with the tags of web images by modeling  $p(\widetilde{y} | R^x_i ), i=1,2,\dots,K$ where $x$ is a web image with label $\widetilde{y}$ and $ \{R^{x}_{i}\}_{i=1}^K $ refers to the set of detected regions of $x$.
$R^{x}_{i}$ is one of the regions, $K$ is the number of regions. 
Second, we have the following two considerations for tags of web images: 
(1) whether target objects appear in the web image (\eg, a web image used for dog classification contains dogs) and 
(2) whether the label is consistent with the object present in the web image (\eg, the breed of dog in a web image is the same as the label of this image).
To eliminate label bias, we couple the two considerations with the detected object proposals of web images, so that we can select object proposals of web images accurately. 

Fig.~\ref{fig:pip} shows the framework of our method, in which the web dataset passes through the region proposal network and the constraints, and then becomes $\mathcal{D}_{w^*}$.
(a) Training of region proposal networks has two-levels of parameters to control the size of bboxes corresponding to the red (region proposal) and blue (object detection) rectangles on the images.
During generation of the proposals on the web dataset $\mathcal{D}_{w}$ (see (b) Prediction of RPNs), there is a two-stage selection based on the above two RPNs trained on bboxes of two sizes.
Meanwhile, the basic model will predict the probability $p(y|R_i^x)$ of proposals, which can be used to remove noisy regions.  
Finally, we take the output $\mathcal{D}_{w^*}$ and  $\mathcal{D}_{s}$ as the training set to train the classification CNN model.
The RPN and constraints are introduced as follows.

\subsection{Unsupervised Region Proposal Networks}
\label{sec:GOB}
As discussed above, candidate object proposals of web images are not only crucial for reducing form bias but also helpful to select regions consisting of relevant tags from web images.
However, since there are no ground-truth annotations to locate object regions, we present a new unsupervised object localization method through joint region proposal and feature extraction via CNN layers without needing bounding box annotations. 
Given a standard dataset $\mathcal{D}_s=\{(x^{(m)},y^{(m)})\}_{m=1...M}$, where $M$ is the number of images, based on the assumption that standard images typically have one object in the center, we define the weak bounding box of an image $x^{(m)}$ as follows: 
\begin{equation}
(l^{(x)}_m,l^{(y)}_m)=(\frac{(1-\lambda)w_m}{2},\frac{(1-\lambda)h_m}{2}),
\label{equ:(1)}
\end{equation}
where $w_m$ and $h_m$ are the width and height of $x^{(m)}$, and $w^{(b)}_m$ and $h^{(b)}_m$ are the width and height of its weak bounding box, which are defined as $w^{(b)}_m=\lambda w_m$ and $h^{(b)}_m=\lambda h_m$.
$(l^{(x)}_m,l^{(y)}_m)$ are the coordinates of the top-left corner of the bounding box,
and $\lambda$ is the proportion used to set the size of bounding box.

We use the standard dataset and the weak bounding boxes defined above to train a region proposal network (RPN) where $\lambda$ is set to $\lambda_{r}$ for region proposal~\cite{ren2015faster}.
Fig.~\ref{fig:pip} shows the framework of the unsupervised region proposal network.
Applying this RPN to web images generates region proposals.
%by using RPN trained on the standard datasets.
%
Given an image $x$, the RPN will output a set of rectangular object proposals  $R^x=\{R_1^x,R_2^x,\cdots,R_K^x\}$, each with an objectness score $p(z|R_i^x)$ that estimates the probability of $R_i^x$ containing a domain object.
The regions produced by the RPN are not accurate enough because of the imprecise bounding boxes used for training.
This is however sufficient, as our goal is to generate candidate regions containing objects from web images rather than locating objects precisely.

To further select appropriate proposals, we use the size ratio $\lambda_{o}$ for training the object detection RPN, which is smaller  than $\lambda_{r}$ used to train the RPN for generating region proposals (as shown in (a) Training of RPNs in Fig.~\ref{fig:pip}).
We train two RPNs by using two different sizes of bboxes for generation of content (red border) and object (blue border) region proposals respectively.

So far we have described how to get a set of candidate proposals which almost contain objects located close to the center.
This makes the scale of related content and the density of domain information similar to the standard dataset. 
During the training of RPNs, one mini-batch is formed by 256 proposals and the loss is calculated based on classification loss and regression loss, which ensure the proposal contains the object, and its location and size are close to the ground truth.

\subsection{Form and Label Constraints}
\label{sec:constraints}
The detection network maps the whole image to domain regions with associated probabilities, but the category of the object in the image has not yet been considered.
Therefore, label bias still exists.
We handle it with convolutional layers followed by a softmax layer.
The convolutional features for detection and classification are shared, so that the classification network can be easily trained.

Given an image $x$, let $R^x=\{R_1^x,R_2^x,\cdots,R_K^x\} $ be the locations of $K$ proposal regions in the image.
% 		%
Note that we have adopted non-maximum suppression (NMS)~\cite{ren2015faster} on all regions to reduce redundancy.
In order to select the distinguishing regions $R_{sub}^x \subset R^x$ from the proposals, we consider combining two score constraints by solving the following optimization problem:
\begin{equation}
R_{sub}^x=\{R^x_i |  \Delta(R^x_i) =1 , i=1, \cdots , K\}, \
\label{equ:optimzie}
\end{equation}
where $\Delta(R^x_i)$ is defined as a scoring function over two constraints as follows:
\begin{equation}
\Delta(R^x_i)=\Delta_{form}(R^x_i)\Delta_{label}(R^x_i),
\label{equ:constraints}
\end{equation}
in which $\Delta_{form}(R^x_i)$ denotes the form constraint and $\Delta_{label}(R^x_i)$ denotes the label constraint,

\textbf{Form constraint.} 
The form constraint controls the scale of the regions. We want to constrain the scale of objects such that they are not ``too small" compared to the size of the regions.
%
%For identifying the object in a region, we train another RPN for object detection with a smaller $\lambda$ than the one used in the region detection model. 
%%%YKL The description here may confuse the reader as it sounds another RPN in addition to the previous two is trained, which is not true. Check the new wording.
%%%XXS, that is  accurate and clear, thanks.
For identifying the object in a region, as described before, in addition to training an RPN with size ratio $\lambda_{r}$ for regions containing relevant content, we also train another RPN with size ratio $\lambda_{o} < \lambda_{r}$ for object detection. 
We then send each region proposal generated by the RPN model with $\lambda_{r}$ 
%%%YKL by the first (object proposal) RPN?
into the object detection model (trained with size ratio $\lambda_{o}$ )
%%%YKL Is this the object detection RPN?
%%%XXS Yep,I have change the symbol to \lambda_{r}$ and \lambda_{o}$ to make it clear.
and retain the proposal with maximum object score as the object proposal. 
Hence, we define the constraint as:
%
%\begin{equation}
%\Delta_{form}(R)=\prod_{i=1}^{k} c(R^x_i)
%\end{equation} 
%%
%where the term $c(R^x_i)$ is
%
\begin{equation}
\Delta_{form}(R^x_i)=\left\{\begin{matrix}
1& IoU(R_i^x) \geqslant \eta \\ %0.7
0& \text{otherwise}
\end{matrix}\right.
\label{eq:form}
\end{equation}
where $IoU$ means the Intersection-over-Union overlap between the object proposal and the region. 
$\eta$ is the threshold value of IoU.
The proposal objects are generalized by the detection model trained on the standard dataset.

\textbf{Label constraint.}
The label of web images may contain errors, so we select proposals by adding a label constraint:
% 
%\begin{equation}
%\Delta_{label}(R)=\sum_{i=1}^{k} c_l(R^x_i)
%\end{equation}
%Training 
%where $c_l$  is the  constraint on label, defined as
%
\begin{equation}
\label{eq:label}
\Delta_{label}(R^x_i)=\left\{\begin{matrix}
1 & \widetilde{y}= \widehat{y}, p(z|R_i^x)\geqslant\epsilon \\ %0.8
0& \text{otherwise}
\end{matrix}\right.
\end{equation}
\begin{equation}
\widehat{y}= \arg \ \max_{y_j} p(y_j|R_i^x)
\end{equation}
where  $p(y_j|R_i^x)$ is the probability of region $R_i^x$ belonging to the category $j$, $j=1,...,L$, which is predicted by the benchmark model (fine-tuned on the standard dataset).
$\epsilon$ is the threshold value to control the score of the detected objects.
$\widehat{y}$ is the predicted label of the image region.
%%%YKL I changed 'image' to 'image region' above. Please check.
%%%XXS have checked, 
%
$p(z|R_i^x)$ is the probability that $R_i^x$ contains the classification content.
The predicted label of the region is required to be the same as the web label $\widetilde{y}$.

\renewcommand{\algorithmicrequire}{\textbf{Input:}}
\renewcommand{\algorithmicensure}{\textbf{Output:}}
\begin{algorithm}[t]
\caption{Unsupervised Detection for Web Data}
\label{alg:IL}
\begin{algorithmic}[1] %这个1 表示每一行都显示数字
	\REQUIRE ~~\\%算法的输入参数：Input
	Standard dataset: $\mathcal{D}_s=\{(x^{(m)},y^{(m)})\}_{m=1...M}$; \\
	Web dataset: $\mathcal{D}_w=\left \{(x^{(1)},\widetilde{y}^{(1)}),\cdots,(x^{(N)},\widetilde{y}^{(N)})\right \}$; \\
	The initialized network model $\mathcal{M}_{pre}: f(x;\bm{\theta}_{pre})\in \mathbb{R}^C $.\\
	
	\STATE Define the weak bounding boxes of images in $\mathcal{D}_s$ by Eq.~\ref{equ:(1)};\\
	\STATE Train an RPN using the weak bounding boxes of images in $\mathcal{D}_s$;\\
	\STATE Generate $R^x$ for $x$ of $\mathcal{D}_w$;\\
	\STATE 	Predict $p(z|R_i^x)$ and probability $p(y|R_i^x)$  of region;\\		
	\FOR {$R_i^x \in R^x$}  
	\STATE  calculate the $\Delta_{form}(R^x_i)$ by Eq.~\ref{eq:form};
	\STATE  calculate the $\Delta_{label}(R^x_i)$ by Eq.~\ref{eq:label};
	\STATE  get $\Delta(R^x_i)$ by Eq.~\ref{equ:constraints};
	\STATE  optimize $R_{sub}^x$ according to Eq.~\ref{equ:optimzie};
	\ENDFOR 
	\STATE $\mathcal{D}_{w*}=\{R_{sub}\}$;
	\STATE Use $\mathcal{D}_{w*}$ to aid the training of CNN model;\\  
	% $\mathcal{M}_{t-1}$ is roughly the same.
	%		$\widetilde{\mathcal{D}}_{t-1} \cap \widetilde{\mathcal{D}}_t \approx \widetilde{\mathcal{D}}_{t-1} \cup\widetilde{\mathcal{D}}_t$
	\ENSURE CNN model trained on  $\mathcal{D}_{w*}$ and  $\mathcal{D}_{s}$ \\	
\end{algorithmic}	
\end{algorithm}

Finally, for each web image, we obtain  $R^{x}_{sub}$ which is de-biased with the standard data. 
%%%YKL I suggest changing unbiased to de-biased as the bias is not entirely removed
%%%XXS that is a good suggestion, I have changed unbiased in other places.
That is, by combining scale and label information, we map  $\mathcal{D}_w$ to a new processed dataset $\mathcal{D}_{w*}=\{R_{sub}\}$, which is used to assist the target task.
%
%%%YKL I add a subheading here as the following does not belong to Form and Label Constraints
%%%XXS Got it.
\subsection{Training Process}

We use $\mathcal{D}_{w*}=\{R_{sub}\}$ as the auxiliary data and input it along with $\mathcal{D}_s$ into the CNN model.
The parameters $\theta$ of the model $\mathcal{M}$ are initialized on parameters $\theta_{pre}$ of the pre-trained model $\mathcal{M}_{pre}$ (trained on ImageNet), and then the parameters are updated using stochastic gradient descent (SGD).
The iteration of SGD updates current parameters 
$\theta^*$ as:

\begin{equation}
\theta^*=\theta+\widehat{\gamma}\cdot \frac{1}{\left | \mathcal{D}^b\right|}\sum_{x,y\in \mathcal{D}^b}\bigtriangledown_{\theta}\left [l(x,y) \right ],
\end{equation}
where $l(x,y)$ is a loss function, \eg, softmax. $\bigtriangledown_{\theta}$ is computed by gradient back-propagation. $\mathcal{D}^b$ is a mini-batch randomly drawn from the mixed training dataset $\mathcal{D}$, and $\widehat{\gamma}$ is the learning rate, which will reduce during training.
The method is summarized in Algorithm~\ref{alg:IL}.
\section{Experiments}

\subsection{Experimental Setup}
\label{sect:ES}
\textbf{Dataset.} 
We employ three standard datasets (where ``standard'' refers to public datasets of the target task), namely Food-101, Stanford Dogs and MIT Indoor67 to evaluate our method. 
We use these datasets for two reasons: 
1) The diverse characteristics of both web-derived and well-labeled datasets, such as the number of instances in each image, object size and location, result in a significant gap which is under-researched in the literature.
2) The representative capacity to various classification tasks, i.e., foreground-oriented shape-based task (dog), foreground-oriented non-shape task (food) and hybrid content based task (indoor scene), can evaluate the robustness of the proposed method.

For each task, the web datasets are collected by keyword search from Google Images, Flickr and Twitter~\cite{yang2018recognition}. 
After downloading web images, the images that are near duplication with any image in the test set are removed  following~\cite{wang2014learning}, which ensures the fairness of our tests.
Finally, 214,743 images of food, 11,2018 images of dog and 76,907 indoor scene web images are used for experiments based on the dataset of work~\cite{yang2018recognition}.
For each source, we use a similar amount of data as the well-labeled images for the food and dog datasets.
%
%After that, we remove those images which are near repeat with the images in the test set (based on the method mentioned in~\cite{krause2016unreasonable} ), to ensure fairness of our test results.
%
For indoor scenes, we found the downloaded web data are more complex (\eg, the contents of web images are richer and there are more similar images) than the other two tasks, so we use more web images (about 2 times compared to standard data) from each source of this task. 
%%%YKL for each source or not?
%%%XXS Yes, I have add several words to make it clear.

\textbf{Implementation.}
The major model of the experiments is the ResNet~\cite{he2015deep}, which has good performance for many classification tasks.
For generating object proposals, we use our data to train the architecture used in~\cite{ren2017object}.
For classification,
the first step of our experiments is to fine-tune CNN models on the target datasets as the basic models.
We use Caffe~\cite{jia2014caffe} in our experiments, and our models are trained on NVIDIA TITAN X GPUs.  
We use different experimental settings on the three datasets because the scales of datasets are different.
The learning rate is initially set to $0.001$, which is divided by 10 after 10 epochs. 
The total number of iterations is 30 epochs with a mini-batch including 20 224$\times$224 images on Titan X (for ResNet110, the batch-size is set to 13 for Titan X).
Since our goal is to obtain content regions related to the label rather than locate objects precisely, the settings of $\lambda_{r}$ and $\lambda_{o}$ for region detection and object detection do not need to be precise, as discussed in Section~\ref{sec:params}.
The detailed discussion about $\eta$ in Eq.~\ref{eq:form} and $\epsilon$ in Eq.~\ref{eq:label} are also shown in Section~\ref{sec:params}.
None of these parameters will seriously affect the results unless they are set to very small or very big values.

\subsection{Parameters}\label{sec:params}
\begin{figure}[t]
\begin{center}
%\fbox{\rule{0pt}{1.5in} \rule{.9\linewidth} {0pt}}
\includegraphics[width=0.9\linewidth]{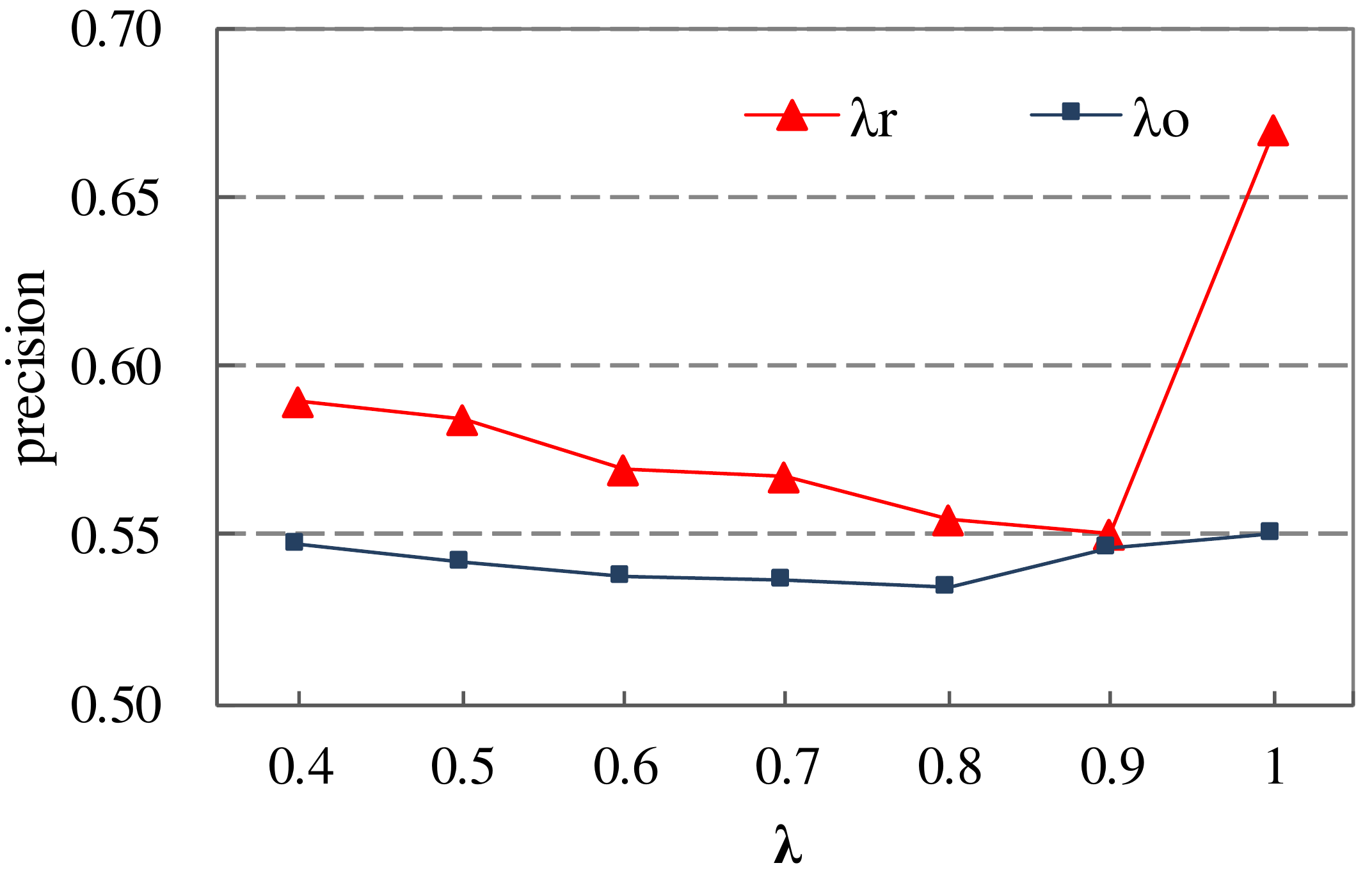}
\end{center}
\vspace{-3mm}
\caption{Precision of ``Name That Dataset'' for classifying Food-101 and de-biased web images, with respect to changing parameters $\lambda_{r}$ and $\lambda_{o}$. 
These values are used for region detection and object detection, respectively.
$\lambda=1$ for region detection means using the pre-trained RPN model to generate regions. \label{fig:lamda}}
%%%YKL text in the figure should be changed to be consistent
%%%XXS it has been changed
\vspace{-2mm}
\end{figure}

\begin{figure}[t]
\begin{center}
%\fbox{\rule{0pt}{1.5in} \rule{.7\linewidth} {0pt}}
\includegraphics[width=0.9\linewidth]{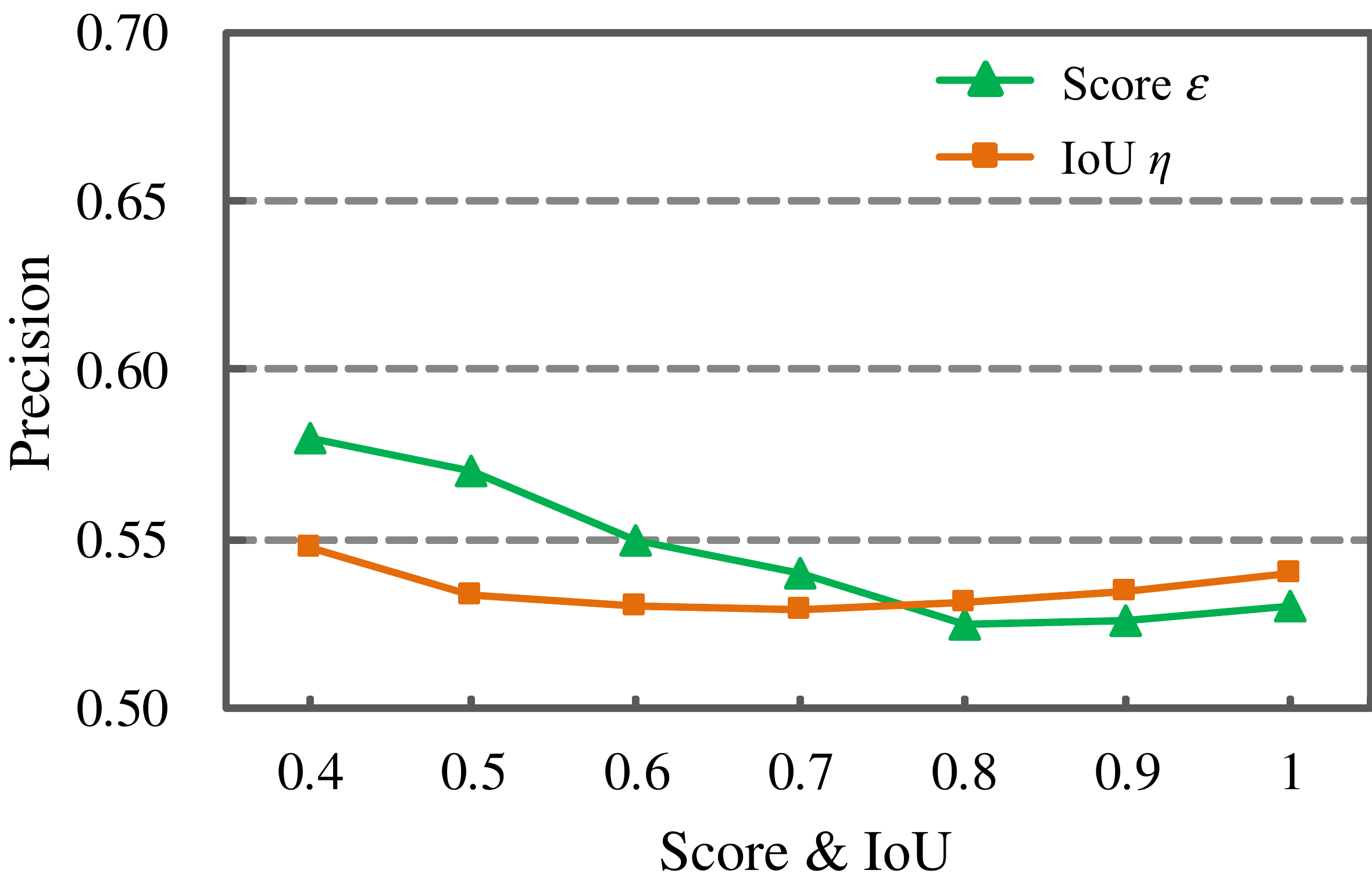}
\end{center}
\vspace{-3mm}
\caption{Precision of ``Name That Dataset" for classifying Food-101 and de-biased web images, with respect to changing parameters: IoU threshold $\eta$ and objectness score threshold $\epsilon$. 
%
%Score $=1$ and IoU $=1$ are approximate values.  
%%%YKL How are they approximate?
%%%XXS that means the two values are similar. I have remove this sentences to prevent ambiguity .
\label{fig:para2}}
\vspace{-3mm}
\end{figure}

In this section, we discuss the parameter settings of Eq.~\ref{equ:(1)}, Eq.~\ref{eq:form} and Eq.~\ref{eq:label}.
Here, we use ``Name That Dataset'' to evaluate how the parameters affect the gap between Food-101 and the processed data.
First, there are two $\lambda$ values for region detection and object detection, respectively.
Fig.~\ref{fig:lamda} shows the classification precision of ``Name That Dataset" on Food-101 and Food-debiased when setting $\lambda_{r}$ and $\lambda_{o}$ to different values. 
As can be seen, $\lambda_{r}=0.9$ is the best setting for region detection, because the minimum of binary classification results is at $\lambda_{r}=0.9$, which means the gap is smaller than other settings.
To make the experiment manageable, we fixed $\lambda_{r}$ for region detection to $0.9$ when discussing $\lambda_{o}$ for object detection, and $\lambda_{o}=0.8$ gets minimum results for classifying the two datasets.
Second, Fig.~\ref{fig:para2} shows the influence of setting the IoU threshold $\eta$ in Eq.~\ref{eq:form} for generating object proposals and the object score threshold $\epsilon$ in Eq.~\ref{eq:label}.

The classification precision of ``Name That Dataset" on Food-101 and Food-debiased with different parameter settings are reported. 
It can be seen that none of these parameters affect the results significantly unless they are set to very small or very big values.
The noise level affects the number of images fed into the CNN model selected by our label constraint module.
For some images, the proposed method can retain more than one region proposal for one image. 
During the training of RPN, one mini-batch is formed by 256 proposals and the loss is calculated based on classification loss and regression loss, which ensure the proposal contains the object, and its location and size are close to the ground truth. 
%
%0.9$\times$ original height/width will not affect the location/size of the proposal since candidates will not be fixed during testing.  
%

\subsection{Evaluation of Built-in Gap Between Datasets}
\label{(sec:bias)}
To analyze the change of the bias between web and standard datasets, we will measure dataset bias from five aspects: (1)\textit{\textbf{ Relative data bias}}: following~\cite{torralba2011unbiased,khosla2012undoing} we define the relative data bias as the uniformity between web and standard data, \ie, if they are misaligned.
%
%\item  %
(2) \textit{\textbf{Cross-dataset generalization}}: similar to the definition proposed in~\cite{torralba2011unbiased,khosla2012undoing}, we use it to evaluate the generalization of different types of data, especially web data.
(3) \textit{\textbf{Scale of related content}}: it is designed based on~\cite{herranz2016scene}, %and shows the characteristics of web data.
and originally means the scale of objects in an image, but in this paper, it represents the scale of the object relevant to the web label.
%
%	\item  
%
(4) \textit{\textbf{Density of domain information}}: it originally indicates the density of objects in a scene image, but here we use this to represent the number of classification objects occurring in a web image.
(5) \textit{\textbf{Label quality}}: it is determined by the noisy label of web data and represents the relationship between web label and web image content. 
%\end{itemize}
%Before processing, 
Specifically, we measure the dataset bias as follows: 

\begin{figure}[t]
\begin{center}
	%\fbox{\rule{0pt}{2in} \rule{.9\linewidth}{0pt}}
	\includegraphics[width=0.9\linewidth]{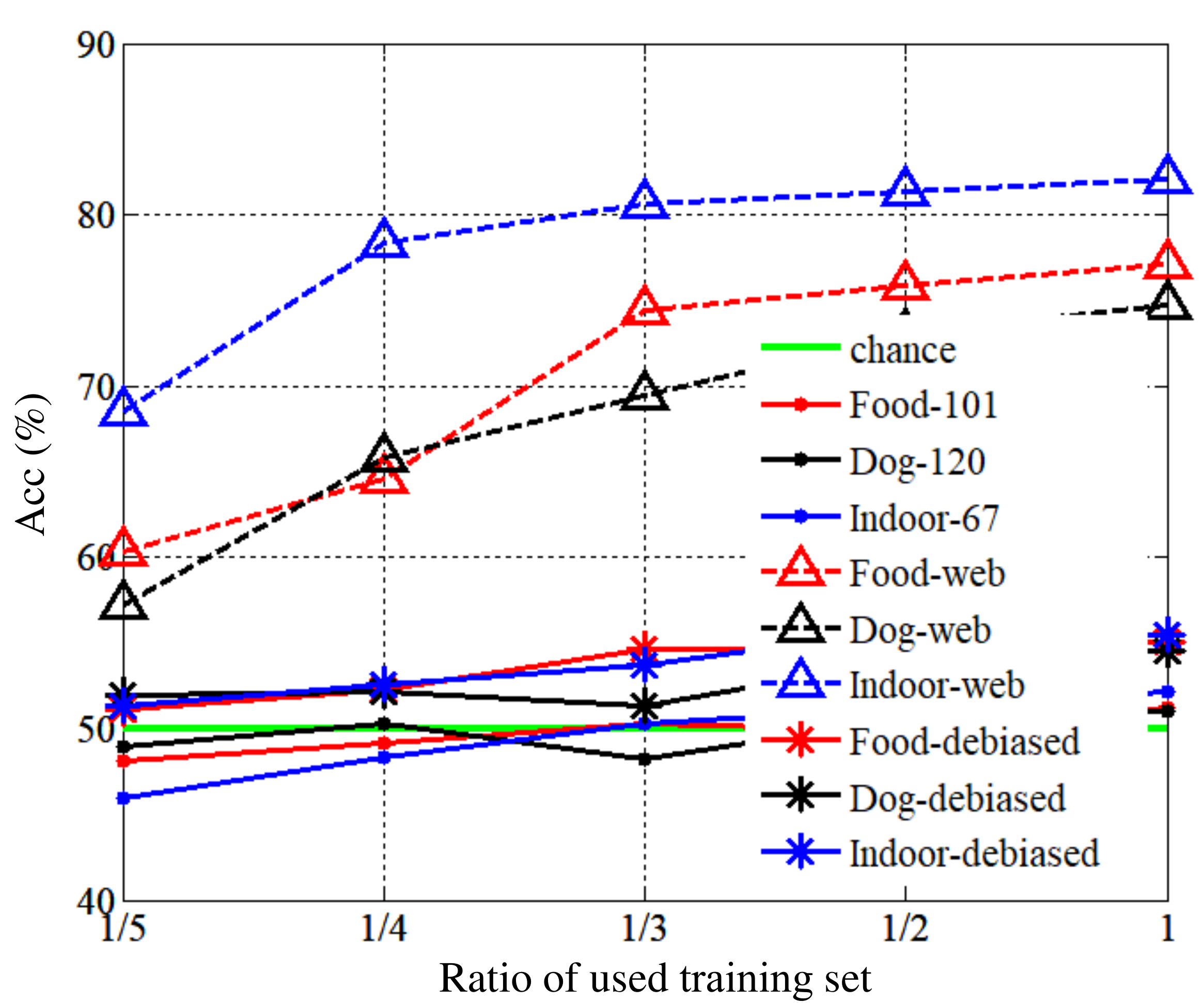}
\end{center}
\vspace{-3mm}
\caption{CNN plays ``Name That Dataset'', where the task is to separate images from standard datasets and web datasets as specified by each curve. The `-debiased'
%%%YKL Change -unbiased in the figure to -debiased
%%%XXS changed.
means to classify standard data and data obtained using our method. ``Ratio of used training set'' is the proportion of training dataset actually used for training. }
\label{fig:bias2}
\vspace{-3mm}
\end{figure}

%\begin{figure}[!t]
%	\begin{center}
%		%\fbox{\rule{0pt}{2in} \rule{.9\linewidth}{0pt}}
%		\includegraphics[width=0.9\linewidth]{bias-food.pdf}
%	\end{center}
%	\vspace{-3mm}
%	\caption{The CNN plays ``Name That Dataset''. Left: example images from Food-101 and the Web.  Right: classification	performance as a function of the amount of training data for three datasets. The performance does not appear to completely saturate. ``-Mix'' means the classification task is to recognize both standard and web data.  ``Acc'' is classification accuracy. ``Ratio of used data'' is the proportion of data used in ``Name That Dataset'' compared to the total dataset. }
%	\label{fig:fig1b}
%\end{figure}

\subsubsection{Measuring Dataset Bias}
In this section, we will report the quantitative evaluation to demonstrate the existence of bias.

First, we employ the game ``Name That Dataset" used in~\cite{torralba2011unbiased} on three datasets respectively as the measure of \textbf{\textit{relative data bias}}. 
As shown in Fig.~\ref{fig:bias2}, for each dataset (Food-101, Dog-120 and Indoor-67), we create a mixed dataset (indicated by suffix -mix) by combining the same number of training images from the training dataset and web images, and we gradually increase the number of dataset images from $1/5$ to the full set.
This is used to train a CNN model for binary classification of whether the image is from the standard dataset or the web. 
With increasing training data, the web and standard images are easily classified by the model. 
For comparison, we also conduct the experiment on each standard dataset itself by labeling half of the images as 1 and the rest as 2, and the classification accuracy is stable at around $50\%$.
The line chart illustrates that web and standard datasets have strong relative data bias and are not aligned, \ie, either of them is unique and identifiable from the other.
%each dataset possesses a unique, identifiable “signature”.
%

Second, the \textbf{\textit{cross-dataset generalization}} experiments on web and standard datasets are designed following~\cite{torralba2011unbiased}. 
%
%We think that collected web data for a target task should represent the information of the target domain. 
%%
%Therefore, we measure the effectiveness of web data.
%
Table~\ref{tab:accb1} shows the classification performance on food, dog and indoor scene datasets and the corresponding web data.
%
%Note that the number of web image for each class used in cross-dataset generalization experiment is same with standard datasets.
Note that the same number of web images as the standard datasets are used in this experiment.
%
%Each row corresponds to the performance of training on the one dataset then testing on its testing set and the other dataset, \eg  in line 1, $84.32\%$ is got by training on Standard(Food-101) and testing on Standard (Food-101), and  $52.49\%$ is got by training on Standard (Food-101) and testing on Web (Food).}
%
\begin{table}[t] \small
\begin{center}
	\caption{Cross-dataset generalization:
		Classification accuracies when training on one dataset (row) and testing on Standard and Web datasets as well as their mean ($\pm$ difference). }	\label{tab:accb1}
	\setlength{\tabcolsep}{2.5mm}{	\begin{tabular}{l| l | l | l }
		\bottomrule
		%\backslashbox{Train}{Test}
		\diagbox[width=9em,trim=l]{Training}{Testing} & Standard &Web& \multicolumn{1}{c} {Mean}  \\ \midrule
		Standard (Food-101)                   & \textbf{84.31}   & 52.49   &$68.40\pm15.91$  \\
		Web (Food)                 & 74.01   &\textbf{81.72}    &$77.87\pm7.71$    \\ 
		\midrule			
		Standard (Dog-120 )                  & \textbf{81.26}  &   58.41    & $69.84\pm11.42$  \\
		Web (Dog)                   &  67.58  & \textbf{74.45} &$71.02\pm3.43$ \\ 
		\midrule			
		Standard (Dog-120)                   & \textbf{81.26}   &   62.79     & $72.11\pm9.15$ \\
		Web	(L-Dog~\cite{krause2016unreasonable})          & 68.09 &  \textbf{73.36}    &  $70.73\pm2.63$ \\ 
		
		\midrule
		Standard (Indoor-67)                & \textbf{79.64} &   60.40  &   $70.02\pm9.62$  \\
		Web (Indoor)           & 65.97  &   \textbf{72.39}  & $69.18\pm3.21$\\ 
		\bottomrule			
	\end{tabular}}
\end{center}
\end{table}
Results in each row are trained using the specified dataset, and the three columns show accuracies when testing on the standard and web datasets as well as the mean of the two accuracies, \eg, in the first row, $84.31\%$ accuracy is obtained by training and testing on Standard (Food-101), and $52.49\%$ is obtained by training on Standard (Food-101) and testing on Web (Food).
Furthermore, to prove that web dataset bias is universal rather than caused by the collection method used in our work, we conduct the same experiment using web data from L-Dog collected by Krause \etal~\cite{krause2016unreasonable}.
%containing 342632 images from 515 categories, in which 120 dog classes are same with Stanford Dogs and 395 are other categories. In cross-dataset generalization experiments, 
In this experiment, we use 21,827 images (13,158 for training, 8,669 for testing) from the L-Dog dataset with the same classes as Stanford Dogs for a fair comparison. 
%

%
%The cross-dataset generalization is $train_s={80.51, 77.32}$, for the data of this work is reviewed by people, so the scened value is not very small, but the distinction still exists.
%
%
As discussed in~\cite{torralba2011unbiased}, the actual accuracies are not important, but the differences in performance are worth discussing.
The performance when trained on one source and tested on the other source is substantially worse than training and testing on the same source (bold), so the two sources are not generalizable to each other.
Meanwhile, the mean $\pm$ difference shows that web data is more generalizable than standard data, which conforms to the fact that standard datasets have more restrictions than web data.

\begin{figure*}[t]
\begin{center}
	%\fbox{\rule{0pt}{6in} \rule{.9\linewidth}{0pt}}
	\includegraphics[width=\linewidth]{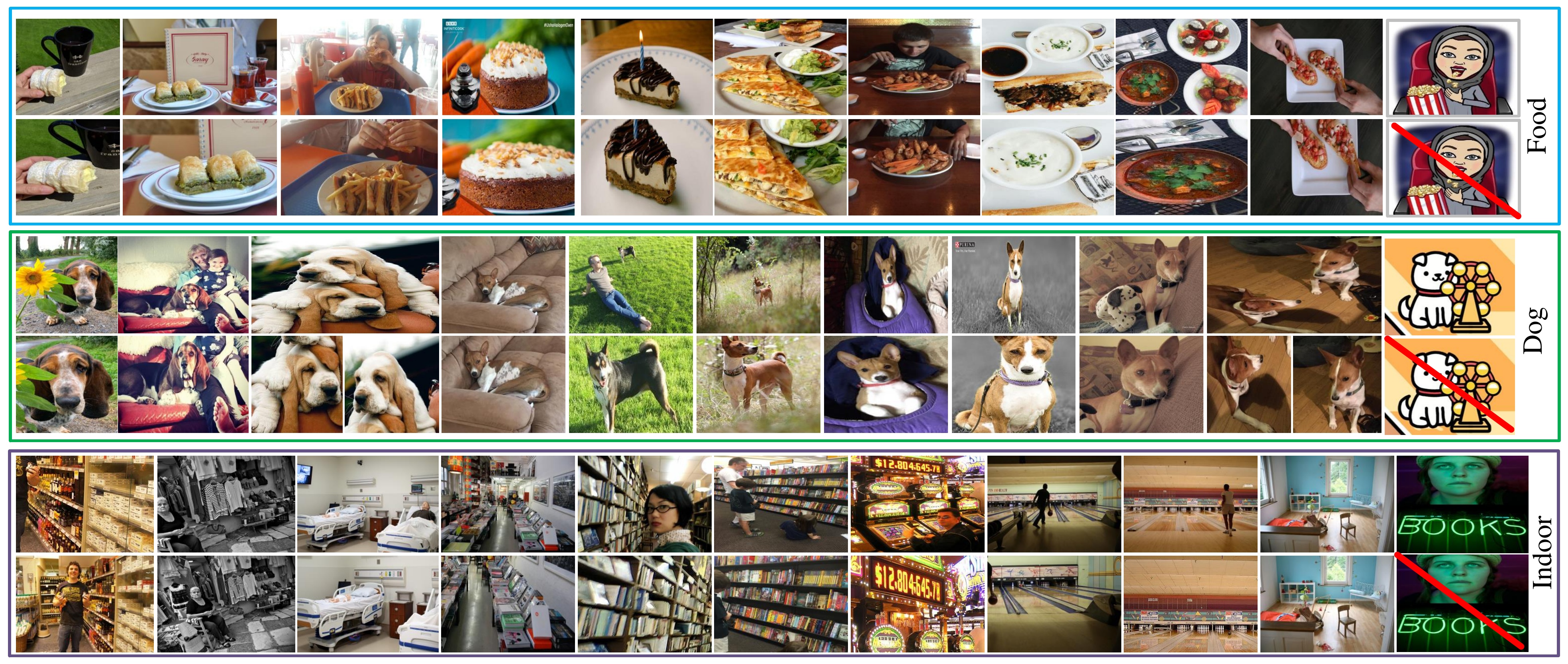}
\end{center}
\caption{Examples of the working mechanism of unsupervised object localization.
We show examples of web images and the corresponding de-biased images. 
	The web images contain the form bias and label bias,
	and there are many images with a density greater than one and our method handles them well.}
\label{fig:eg}   
\vspace{-3mm}
\end{figure*}
Third, we further perform two experiments to measure the \textbf{\textit{scale of related content}} and \textbf{\textit{density of domain information}}. 
The scale of related content measures the size of the detected region containing the relevant content compared to the size of the entire image, and the density of domain information is the number of detected regions related to the subject in the image.
The scale is defined as
\begin{equation} \textrm{scale}=\frac{1}{d}\sum_{i=1}^{d}\frac{|I_i|}{|I|},
\end{equation}
where $d$ is the number of detected regions related to the subject of the domain task in the image, so the density is $d$.
$|I_i|$, $i=1 \dots d$ is the size of the $i$-th region containing relevant objects, and $|I|$ is the size of the image.
%
%Fig.~\ref{fig:eg} shows images randomly picked from standard dataset and Web dataset. 
%
In Table~\ref{tab:scale}, we report the average scales and densities of images on Food-101 and Stanford Dogs datasets.
We can see that the densities of standard images are almost always 1, but they are getting bigger on web images.
Meanwhile, the scale of related content is usually smaller in web images than in standard images.
For example, Food-101 has $density=1.16$ and $scale=0.8536$ whereas Food-web has $density=1.94$ and $scale=0.6218$, which are distinguished from each other.
Different from object classification, indoor scene recognition requires knowledge about both scenes and objects.
Calculating the scale and density of a scene image is not reasonable, but we can take a web scene image as a combination of several
%%%YKL Not clear if these metrics are used for indoor scenes or not?
%%%XXS during experiments we used the same process, but we do not use the metrics to calculate the changes before and after process(as table 2), because the detected region on scene image not only dependent on object and many images do not have detected region. We add one sentence to illustrate.
parts (some contain useful information and some are redundant information \eg, people).
Therefore, we directly use the same process on scene data as for food and dog images without measuring the changes of these metrics.
% 0.1018  0.1447  0.0203  0.0364  0.0517  0.1013 
%
%\begin{figure}[t]
%	\vspace{-2mm}
%	\begin{center}
%		\includegraphics[width=0.9\linewidth]{bias-ex22.pdf}
%		\vspace{-1mm}
%	\end{center}
%	\caption{Illustration of scale and density related with task information. In the bottom boxes, we report the scale and density (scale; density) of Standard (S) and Web (W) datasets for each task. ``Scale'' is the distribution of object sizes and ``Density'' is the number of labeled objects per image. However, Indoor scene requires knowledge about both scenes and various objects, so we set the density to default value 1 and do not control its scale.}
%	\label{fig:fig1bb2}
%	\vspace{-5mm}
%\end{figure}

%
Finally, for the label quality, we use the region proposal model (see Section~\ref{sec:GOB}) to generate content regions for web images and find about $31.3\%$ of webly-labeled food images have no detected regions. 
The result shows that about $1/3$ of webly-labeled food images are outliers.

Overall, the above results demonstrate the existence of dataset bias between web and standard datasets.

\begin{figure*}[t]
\begin{center}
	%\fbox{\rule{0pt}{2in} \rule{.9\linewidth}{0pt}}
	\includegraphics[width=\linewidth]{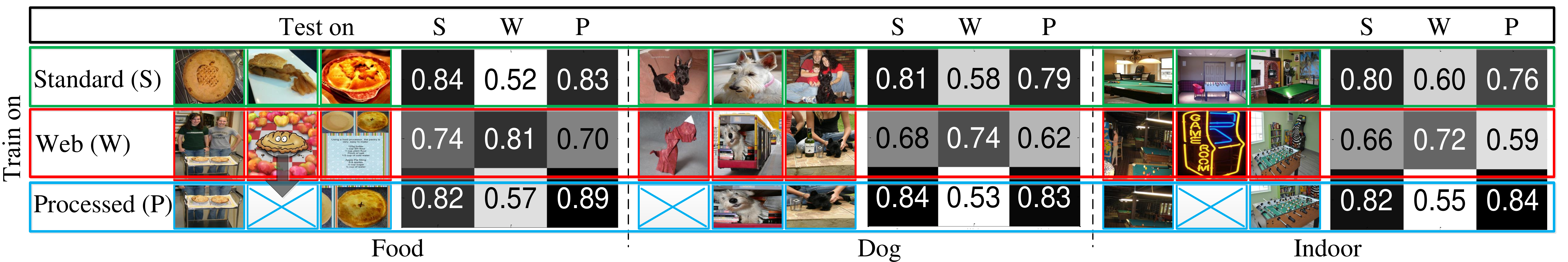}
\end{center}
\vspace{-3mm}
\caption{Cross-dataset generalization. Green, Red and Blue mean Standard, Web and Processed (debiased) web images respectively.
The values are the recognition accuracy of the models that are trained on Standard (S), Web (W) and Processed (P) images respectively, and tested on Standard (S), Web (W) and Processed (P) data (columns).
We also show some example images for each kind of data.
For each column, the image with green border comes from standard dataset,
and the two images in second and third rows are a pair, which are one web image (red) and the corresponding image after processing (blue).}
\label{fig:bias}   
\vspace{-3mm}
\end{figure*}

\subsubsection{Estimating the Culprits of Built-in Gap}
\label{sec:type bias}
We now investigate the causes of built-in gap between web and standard datasets. 
%

%\textbf{Selection of data.} Standard datasets are usually carefully collected, selected and labeled manually, but web images are directly retrieved via searching keywords without additional processing.
%
%Since web images are uploaded by people from different countries with different experience and cultural background and are tagged in a casual way, various labels have been attached to images from the same category. 
%
%In previous work, this bias is one kind of noisy. 
%
%Furthermore, social websites and search engines have different searching mechanisms and topic information, such as for the key words of beer, social websites will feedback images of some occasions to drink beer, but search engines will feedback different kinds of beer.
%%%YKL What do you mean by 'topic information'?
%%%XXS we mean social websites and search engines for different target so the they will returns images of  different subject for the same key words. I have added an examples
%which will also introduce differences to the search results.
% 
%Therefore, different selection processes result in the bias.

\textbf{Forming of datasets.}  It is a more generic concept of ``capture bias''~\cite{torralba2011unbiased} relating to focal length, finder frame and view \etc during shooting --- that objects in images from the standard dataset 
%%%YKL Changed from 'target' to 'standard' above. Please check.
%%%XXS have checked.
are almost always in the center on the image, but objects in web images are not always single-object-in-the-center.
Moreover, as shown in Fig.~\ref{fig:eg}, the contents of web images are rich, not just containing one major object with a big scale. 
These factors result in the form bias.

\textbf{Label noise level of datasets.} It comes from the fact that images of a domain task (fine-grained classification) are often difficult to label for people because of the large intra-class and small inter-class variations among objects. 
This is particularly problematic for web data, which are usually labeled by non-experts without moderation. 
Image labels may be inaccurate or even entirely wrong, \ie, irrelevant content in the image.
These ``wrong'' labels also contribute to the bias.
%
% \textbf{negative set bias}  The negative set defines what the dataset considers to be the other task. If that target set is not representative, or unbalanced, that could produce the CNN model that are overconfident and not very discriminative on the same domain task. 
%
The three kinds of biases are with respect to the information distribution, content and label of web images.
For form and label bias, we show the results of reducing the influence of them as follows.  
%%%YKL The last sentence above appears wrong. This should be revised.
%%%XXS I have rewritten the sentence.
\subsection{Performance of Reducing Bias}
%%%YKL Why call this baseline -- it appears to be different version of the new method?
%%%XXS Because we think the changes of bias is the basic of the method. I have changed it to a new subtitle.
\label{sect:performance}
The effects of the proposed method on reducing bias between web and standard datasets are shown as follows.

In order to show the change of dataset bias quantitatively, we employ the game ``Name That Dataset" again.
We process the web data by our method, and select the same number of images from different sources to do the experiments. 
The results are shown in Fig.~\ref{fig:bias2}, after debiasing, the \textbf{\textit{relative dataset bias}} becomes much lower  (``-debiased'' datasets), comparable to the results on two halves of standard data (around 50\%).
%
%The recondition of ``data-undo" and standard data becomes difficult,
%
%%%YKL some numbers are missing.
%%%XXS -mix is chanegd to -web
Furthermore, with more processed training data (-debiased) added, the classification accuracies keep leveling at around the chance $50\%$.
On the contrary, the classification accuracies of  ``data-web" tend to rise with more training data.
Because the images after processing will be similar to the images from the standard dataset, the bias between web and standard datasets is reduced.
Fig.~\ref{fig:eg} shows some examples of the images before and after processing, in which the scale and density of objects in the images are changed.
Moreover, the noisy images are removed during processing.
%
%--------------------------------------------------------------------------------------
\begin{table}[t] 
\caption{Average scale and density values of standard, web and de-biased web data.
The regions of standard data are from RPN and have correct predicted labels.
For web data, the regions and calculations of scale and density are based on the RPN ($\lambda_r$).
For debiased-web data, regions are from the RPN ($\lambda_r$) trained on the standard dataset and have correct predicted labels, and the scale and density are calculated by the output of RPN with a smaller $\lambda_o$ for de-biased-web data.
%%%YKL The sentence in the brackets is confusing. Clarify which RPN is used in either case.
%%%XXS have modified.
\label{tab:scale}}
\begin{center}
\setlength{\tabcolsep}{8mm}{	\begin{tabular}{l|c |c}
	\toprule
	Dataset                                  & Scale & Density\\ 
	\midrule
	Food-101  & 0.8536  &1.16\\
	Food-web  &0.6218& 1.94 \\
	Food-debiased & 0.7775& 1.23 \\
	\midrule
	Dog-120  &  0.7381 & 1.00 \\
	Dog-web  & 0.5939 & 1.45 \\
	Dog-debiased &0.7465 & 1.18 \\
	\bottomrule
\end{tabular}}
\end{center}
\end{table}
%-------------------------------------------------------------------------------------- 
%
As shown in Fig.~\ref{fig:bias}, the \textbf{\textit{cross-dataset generalization}} results show that de-biased web datasets become significantly more generalizable: training on standard datasets and testing on de-biased web data and vice versa both show significantly better performance than with original web data.

For the \textbf{\textit{scale of related content}} and \textbf{\textit{density of domain information}}, we report the measures before and after bias reduction using our method in Table~\ref{tab:scale}.
Our method simultaneously removes different biases: 
the label constraint in our method can remove noisy web images and the form constraint controls the scale of the related object, and the density of domain information becomes more consistent with the target dataset. 
95,672 (214,743), 68,355 (112,018) and 37,496 (76,970) are the numbers of images after (before) elimination for the three tasks. 
The numbers of proposals are $286,180$, $167,925$ and $153,772$ for web food, dog and indoor scenes, respectively.
After processing, the numbers of the regions are $121,327$, $80,658$ and $72,148$.
\begin{figure}[t]
\begin{center}
  	\includegraphics[width=0.9\linewidth]{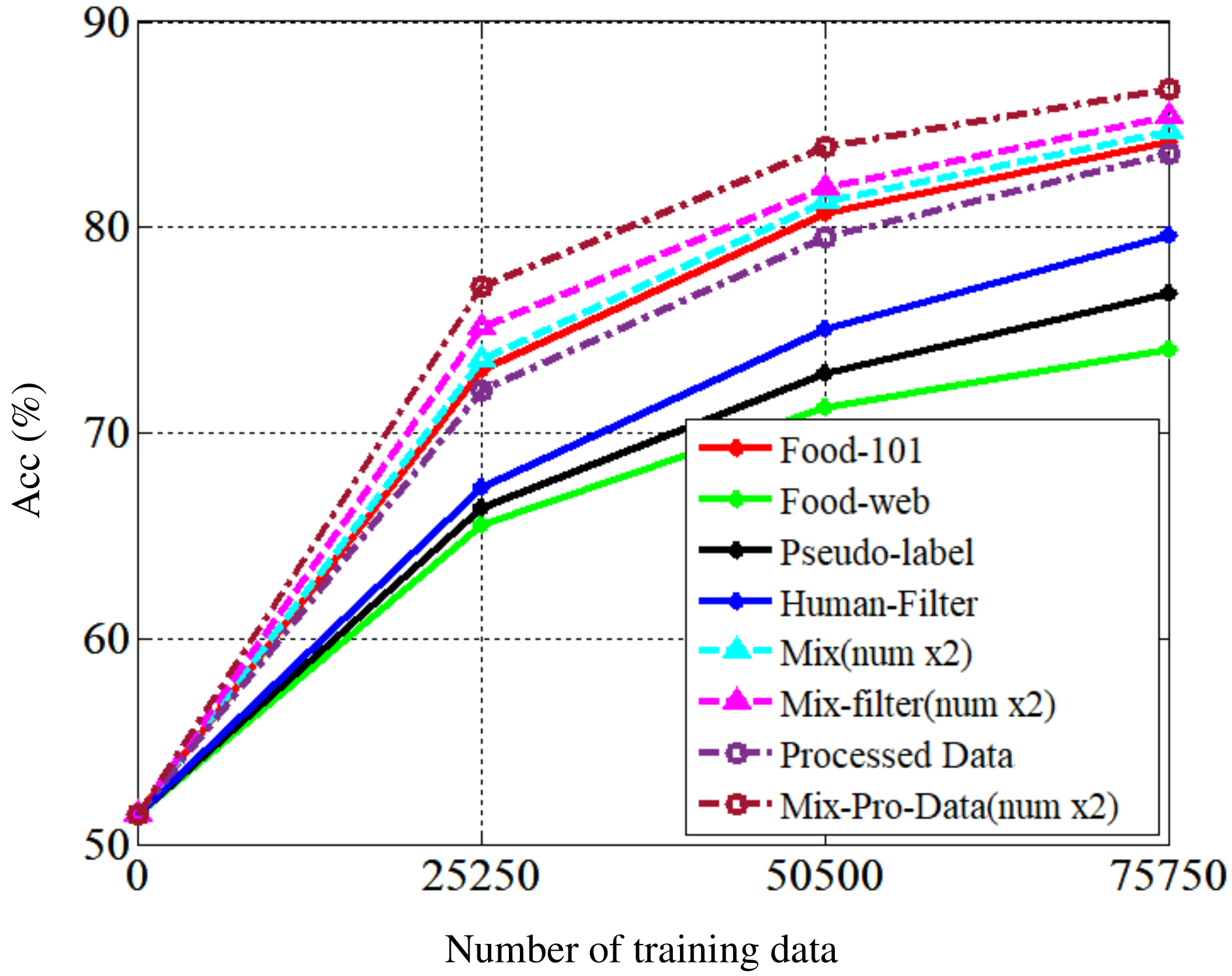}
\end{center}
\caption{Performance comparison demonstrating the gain of adding web data and processed data.
  	``Mix" means the web and standard data are added together, so the number of images is twice of only using Food-101 or Food-web.
  }
\label{fig:add-data}
\end{figure}

To analyze the effect of using web data, we evaluate the performance of models trained on images from the following sources: the Food-101 dataset~\cite{bossard2014food}, web food images (Food-web), as well as web images after manual filtering (Human-Filter) and processed by Pseudo-label method~\cite{lee2013pseudo}.
The test set is the standard set of Food-101. 
%which is taken as the target task with insufficient training data.
%
As shown in Fig.~\ref{fig:add-data}, the improvements of performance are disparate when adding the same amount of standard data (red border) and webly-labeled data (green border).
This may be due to the fact that web data contains noisy labels.
%, for which, the web data contains noise that may be a suitable theoretical explanation so far.
%
\begin{table}[t]
\begin{center}
  	\caption{Performance gain of adding processed web data. Acc means classification accuracy. \label{tab:add=pd}}
  		\setlength{\tabcolsep}{3mm}{\begin{tabular}{l| c c c c}
  		\toprule
  		%	\# unbiased images & 50500 & 63125 & 75750 & 95672\\
  		Amount of Processed data & $25\%$ & $50\%$ & $75\%$ & $100\%$\\
  		\midrule
  		Food-101 Acc (\%)& 85.24 & 86.23 & 88.56 & 90.41\\ 
  		\midrule
  		Stanford Dogs Acc (\%)& 81.20 &83.49 & 85.13 & 87.95\\    
  		\midrule
  		Indoor-67 Acc (\%)& 80.82 & 81.63 & 83.25 & 84.93 \\ 
  		\bottomrule
  	\end{tabular}    }
\end{center}
\end{table}	

\begin{table*}[t] 
\begin{center}
  	\caption{Experimental results on the Food-101 dataset. ``Clean-ft'' means fine-tuning on the model pre-trained on ImageNet with Food-101, and ``Mix-ft''  means fine-tuning with both standard and web data. 
  		Method Accuracy and Our Accuracy are accuracies with the method itself, and the proposed method.
  		Numbers in the brackets show performance gain compared with ``Mix-ft'' as the baseline.
  		\label{acc1}}
  	\setlength{\tabcolsep}{4.5mm}{\begin{tabular}{p{3cm}|p{1.5cm}<{\centering}|c|c|c|c }
  		\toprule
  		\multicolumn{1}{c|}{\multirow{2}*{Method}}  &  \multicolumn{1}{c|}{\multirow{2}*{Model}}        &      \multirow{2}*{\tabincell{c}{Clean-ft \\ Accuracy (\%)}}      &       \multirow{2}*{\tabincell{c}{Mix-ft \\ Accuracy (\%)}}        & \multirow{2}*{Method Accuracy (\%)}  & \multirow{2}*{Our Accuracy (\%)}                   \\
  		& & & & &\\
  		\midrule
  		Bottom-up~\cite{sukhbaatar2014learning} &       & \multirow{3}*{66.49 } & \multirow{3}*{68.72} & 70.29 ~ (+1.57) & \multirow{3}*{\textbf{75.31~ (+6.59)}} \\ 
  		%  \cline{1-1}\cline{5-5}
  		Pseudo-label~\cite{lee2013pseudo}       & AlexNet    &                       &                      & 69.36 ~(+0.64) &  \\ %\cline{1-1}\cline{5-5}
  		Weakly~\cite{joulin2016learning}        &       &                       &                      &  71.10 ~(+4.61)           &  \\ \midrule
  		Boosting~\cite{sukhbaatar2014training}  &      & \multirow{3}*{67.43}  & \multirow{3}*{69.00} & 72.53 ~(+3.53) & \multirow{3}*{\textbf{75.58 ~(+6.58)}} \\
  		% \cline{1-1}\cline{5-5}
  		PGM~\cite{xiao2015learning}             & CaffeNet     &                       &                      & 73.14 ~(+4.14) &  \\ 
  		%\cline{1-1}\cline{5-5}
  		WSL~\cite{chen2015webly}                &      &                       &                      & 73.21 ~(+4.21)  &  \\ 
  		\midrule
  		Harnessing~\cite{vo2017harnessing}      & VGGNet         & 74.32 & 76.98 & 79.02 ~(+2.04) & \textbf{81.93 ~(+4.95)}\\ 
  		\midrule
  		Goldfince~\cite{krause2016unreasonable} & \multirow{2}*{ResNet50} &  \multirow{2}*{84.31}& \multirow{2}*{84.89} & 86.75~ (+2.86)&\multirow{2}*{\textbf{90.41 ~(+5.42)}} \\ 
  	%	\cline{1-1}\cline{5-5}
  	DPTL+subA~\cite{cui2018large} &  &  &  & 88.78 ~(+3.89) &	\\
  		\bottomrule
  	\end{tabular}}
\end{center}
\end{table*}

However, the filtered web data (blue and black) has a much lower noise level, yet the performance is still significantly worse than the standard data (red) of Food-101. 
We also use mixed data (Mix) of web and Food-101 as well as filtered web data (Mix-filter) to train CNNs.
The number of mixed data is double of Food-101 and Food-web for the same $x$ coordinate in the graph, but there is no significant improvement on the results after filtering the web data.
The reason is that the web data and filtered data have poor cross-dataset generalization. 
On the contrary, the results of using the processed data are similar to the results of using the standard data (red).
Furthermore, mixing the processed data with the standard data (Mix-pro-Data) gets improved results than all the other methods. 
For the three tasks, the proposed method achieves consistently increasing performance with more processed images and the results are shown in Table~\ref{tab:add=pd}.
%%%YKL I don't understand the last sentence, do you mean: ... can achieve consistently increasing performance with more processed images?
%
%%%XXS, yes, I have modified this sentence.

%
\begin{table}[t]\scriptsize
\begin{center}
\caption{Experimental results on the Food-101, Stanford Dogs and Indoor-67 datasets.
``ft'' means fine-tuning on the model pre-trained on ImageNet.
``+web" means adding the web data into the standard training set.
``+$\Delta_{form}\Delta_{label}$" represents that debiased web data is used.
``Acc" is classification accuracy. \label{tab:acc-ourmethod}} 
\setlength{\tabcolsep}{2.6mm}{	\begin{tabular}{l| l| c | c | c }
	\toprule			 
\multicolumn{2}{c|}{\multirow{2}{*}{Method}} & \multicolumn{3}{c}{Acc (\%)} \\  
\cline{3-5}
\multicolumn{2}{c|}{} & Food & Dog & Indoor\\	
	\midrule
	\multirow{8}{*}{\rotatebox{90}{Object-Detection}} 	&Grad-CAM~\cite{selvaraju2017grad} &  82.94 & 84.92& 78.55\\
	& Cascade R-CNN~\cite{cai2018cascade} & 82.45 &82.04 & 78.66\\
	&RefineDet~\cite{zhang2018single} &83.20 &82.68 &79.91 \\
	&  DSOD~\cite{shen2017dsod} & 84.75 & 81.93 &78.51\\
 & Cross-Domain WSD~\cite{inoue2018cross} & 84.97 & 83.26&79.43\\
	& ResNet-50+EdgeBox~\cite{zitnick2014edge} &  85.37 & 81.74 & 78.16\\
	& 	ResNet-50+Selective Search~\cite{uijlings2013selective} & 84.82 & 81.43 & 77.69 \\
	& 	ResNet-50+DeepMask I~\cite{pinheiro2015learning} &86.59 &83.31 & 78.94 \\
	& 	ResNet-50+DeepMask II~\cite{pinheiro2016learning} &86.34 & 83.86 &  79.25\\
%	& Bottom-up~\cite{cho2015unsupervised} & 86.32 &  82.91 & 80.20\\
	& Mimic~\cite{li2016image}& 86.70& 84.17 & 76.47\\
	\midrule
	\multirow{4}{*}{\rotatebox{90}{Baseline}} 	&	ResNet-50+ft                               & 84.31 & 81.26 & 79.63  \\
	&	ResNet-110+ft                  & 84.88  & 81.75 & 80.82  \\
	&	ResNet-50+(web)+ft              & 84.89 & 82.72 & 82.31  \\
	&	ResNet-110+(web)+ft             & 85.37 &82.89 & 83.73 \\
	\midrule
	\multirow{4}{*}{\rotatebox{90}{Ablation}}	
	&	ResNet-50+(web+filter)+ft      & 86.10 & 83.41 & 80.26  \\
	&	ResNet-50+(web+RPN)+ft      & 86.64  & 82.65 & 79.52\\
	&	ResNet-50+(web+RPN+$\Delta_{form}$)+ft      & 88.15 & 85.71 & 80.66 \\
	&	ResNet-50+(web+RPN+$\Delta_{label}$)+ft      & 89.32 & 85.24 & 82.09    \\
	
	\midrule
	\multirow{2}{*}{\rotatebox{90}{Ours}} 	& 	ResNet-50+$\Delta_{form}\Delta_{label}$  +ft                  & \textbf{90.41} & \textbf{87.95} & \textbf{84.93}  \\
	& ResNet-110+$\Delta_{form}\Delta_{label}$ +ft                       & \textbf{91.63} & \textbf{88.62} & \textbf{85.22}\\
	\bottomrule
\end{tabular}}
\end{center}
\vspace{-2mm}
\end{table}

\subsection{Training CNN Models for Classification}
\label{sect:IC}
We evaluate our method on three datasets. These datasets contain specific forms of objects (dogs),  objects with irregular shapes (food), and even scenes which are of complex forms. The detailed 
results are as follows.

\begin{table*}[t]
\begin{center}
	\caption{Comparison with image classification methods on Food-101, Stanford Dogs and Indoor-67 datasets.
		``Acc" is classification accuracy. There are also some works using web data, such as Goldfince~\cite{krause2016unreasonable} and Progressive filter~\cite{yang2018recognition}. \label{tab:com-method}} 
	\setlength{\tabcolsep}{4.5mm}{\begin{tabular}{c| l| c |  l| c |  l| c }
		\toprule
		\#	&	Method   on Food-101  & Acc (\%) & Method   on Dog-120  & Acc (\%) & Method   on Indoor-67  & Acc (\%) \\ \midrule
    & Random Forest~\cite{bossard2014food} &50.76 &NAC~\cite{simon2015neural}&68.61 &IFV+DMS~\cite{doersch2013mid}& 66.87\\        &SNN~\cite{SNN}&69.90  & FoF-Weakly~\cite{xu2017friend}  & 71.40 &FB/REF~\cite{vo2017harnessing}& 61.60 \\
		\multirow{8}{*}{\rotatebox{90}{Related work}} 	&	DNNFM~\cite{tatsuma2016food}             & 58.49  &  	PDFS~\cite{Zhang_2016_CVPR}                          & 71.96	 &
		MPP \cite{yoo2015multi}                    & 75.67 	\\
		&	DCNN+ft~\cite{bossard2014food}           & 68.44  
		& 			FB/REF~\cite{vo2017harnessing}         & 73.10 &
		MetaObject-CNN \cite{wu2015harvesting}                        & 78.90 \\
		&	PTFT~\cite{7169816}                      & 70.41   &
       FOAF+ft~\cite{7358138}  &74.49  	&
		SFV+place \cite{dixit2015scene}         & 79.00 \\
		&	Im2Calories~\cite{meyers2015im2calories} & 79.00   &
		Weakly-S\cite{Zhang2016Weakly}                       & 80.43 & Places205-VGG~\cite{zhou2018places}
			& 79.76\\
		&	Inception-v3~\cite{hassannejad2016food}   & 88.28   & ZSL-WL~\cite{niu2018webly}  & 85.16&
	MPP+DSFL \cite{yoo2015multi}   &80.78\\
		&  DLA~\cite{yu2018deep} & 89.70 & Goldfince~\cite{krause2016unreasonable} & 85.90 & 	Double fully hybrid \cite{herranz2016scene}   &80.97\\
 & Progressive filter~\cite{yang2018recognition} & 89.77 & Progressive filter~\cite{yang2018recognition} & 87.36 &  HP-Net~\cite{wang2018towards}& 83.10	\\
 & DPTL~\cite{cui2018large} &	90.40 & DPTL~\cite{cui2018large}
		&  88.00 & Progressive filter~\cite{yang2018recognition} & 84.78 \\
		& Ours  &\textbf{91.63}&  Ours  & \textbf{88.62}&  Ours & \textbf{85.22}  \\
		\bottomrule
	\end{tabular}}
\end{center}
\end{table*}

\textbf{Food-101.} 
It is a dataset with 1,000 images of each class.
The training and testing are split into 3:1. 
The number of web images we used is 214,743.
First, Table~\ref{acc1} shows the comparisons with other works utilizing web images or weakly learning from web labels. 
To compare with these methods, we conduct experiments based on different models used in these works on the Food-101 dataset, \eg, AlexNet and VGGNet \etc.
``Clean-ft Accuracy" is the result of training the deep model on standard dataset
and ``Mix-ft Accuracy" is the performance of the model trained on web and standard datasets together.
Combining such methods with our bias removal significantly improves the performance of classification.
``Method Accuracy" is the result of using the method in the first column for learning from web food images.
``Our Accuracy" indicates the result of our proposed method. 
Compared with other methods using web data, the proposed method can further improve the performance.
Moreover, no matter which deep model (AlexNet, CaffeNet, VGGNet, ResNet) is employed, the proposed algorithm shows its superiority consistently.
%
%It indicates that unbiased web data can improve the performance of CNNs effectively.

%
Furthermore, Table \ref{tab:acc-ourmethod} presents the results on the Food-101 dataset of some detection methods, and the baseline as well as the ablation experiment results on ResNet50 and ResNet110 with web data.
As can be seen, the result of ResNet-50+(web+RPN)+ft is similar to ResNet-50+(web+filter)+ft, because the regions generated by the RPN without the constraints still contain many noisy and redundant regions,
which will influence the final results.
Adding $\Delta_{form}$ helps remove redundant regions, so the result is improved by about 2\%.
Meanwhile, introducing  $\Delta_{label}$  further filters the noisy regions to improve the performance.
The ablation experiment results illustrate that each part of our method is efficient.
Our result is 6.25\%  higher than the performance of directly using web data on ResNet-110.
Meanwhile, we also compare with some existing unsupervised methods, such as~\cite{li2016image} and~\cite{inoue2018cross}.
Since the proposed method considers the characteristics of web images and adds constraints on the object proposals, the final generated regions are more similar to standard data than other localization methods.

\textbf{Stanford Dogs.} 
We also evaluate our method on Stanford Dogs to prove its robustness. 
Stanford Dogs contains 120 dog categories, with 12,000 images for training and 8,580 images for testing. 
The number of web images is 112,018 in this work.
Table~\ref{tab:acc-ourmethod} shows the results of fine-tuning the ResNet on Stanford Dogs without web data achieves an accuracy of $81.26\%$.
After adding web data without any pre-processing, the accuracy is only improved by $1.46\%$.
However, by employing de-biased web data, the performance of the model improves substantially by $6.87\%$.
For the dog dataset, the first round filtering removes almost 40\% web images, which contain many useful images with abundant domain information.

We also conduct an experiment on another web dataset L-Dogs to illustrate the universality of the existence of bias.
The  L-Dogs dataset contains the same 120 classes and is collected from Google Images by Krause \etal~\cite{krause2016unreasonable}.
The result of 85.90\% for Goldfince~\cite{krause2016unreasonable} is achieved by using 342,632 images from 515-category dogs, and our method achieves higher accuracy by using only $1/3$ of data in that work.
%
%Finally, we also conduct experiments on the L-Dog dataset, which is a publicly available noisy dataset for dog recognition.
%
The result tendency of experiments on L-Dogs is consistent with that on our collected web data, \eg 82.59\% for ResNet-110+(L-Dog) and 86.55\% for ResNet-110+(L-Dog)+$\Delta_{form}\Delta_{label}$+ft.

\textbf{MIT indoor67.}
MIT Indoor67 contains 67 categories of indoor images, 5,360 for training and  1,340 for testing.
Scene recognition requires knowledge about both scenes and various objects, making the task more challenging.
As shown in Table~\ref{tab:acc-ourmethod}, the basic performance of ResNet50 is $79.63\%$, and the improvement for adding web data is $2.74\%$, and after removing the bias, the accuracy improves by $5.3\%$.
Since the scale and density of domain information in scene images are more difficult to define than in object images, the improvement is not as significant as the results on object recognition datasets. 
Different from object classification, the intra-class variation is more obvious for scenes, so the web indoor scene images are more complex.
Meanwhile, the results of other object detection methods can also prove that scene images are difficult to locate the labeled information, so more noisy regions (focusing on the object rather than the scene) will be introduced to influence the final results.
Nevertheless, the classification accuracy of the proposed method is still higher than existing methods for scene recognition.
Because the form constraint is a soft constraint for the content of the image, it will not break the wholeness of the scene.

In addition, we show the comparisons of related work on each task in Table~\ref{tab:com-method}.
In these works, Goldfince~\cite{krause2016unreasonable} and DPTL~\cite{cui2018large} also use web data, and the amount of used data is larger than that used in our work.
However, the proposed method gets better results by reducing the bias between the web and standard datasets.
Compared to Progressive filter~\cite{yang2018recognition}, our method can save more time for training, because Progressive filter needs to train the model for several rounds.

The proposed method effectively uses the web data to improve the performance of the classification models on all the three tasks.

\section{Conclusion}
In contrast to previous works, in this paper, we reveal the phenomenon of web dataset biases and carry out rigorous quantitative analysis on them from various aspects. 
By conducting extensive experiments, we demonstrate that dataset bias causes the limited benefits of using web data.
To address this problem, we present an unsupervised object localization method to provide critical insights into the object density and scale.
%
%Meanwhile, we collect web data for food, dog and scene and use them to train CNNs.
%
Experiments show that the proposed method effectively reduces dataset bias.
By employing de-biased web data, our method performs favorably against the state-of-the-art on multiple classification tasks.
Web images are known to be easy and cheap to access. 
Although eliminating bias has not been fully solved, this work shows promising benefits towards this direction.
How to use astronomically large webly-labeled data for a specific target learning task remains an open problem for future investigation.

{\small
\bibliographystyle{IEEEtran}
\bibliography{egbib}
}

\end{document}